\documentclass[10pt,twocolumn,letterpaper]{article}

\PassOptionsToPackage{table}{xcolor}

\usepackage[pagenumbers]{iccv} % To force page numbers, e.g. for an arXiv version

\definecolor{iccvblue}{rgb}{0.21,0.49,0.74}
\usepackage[pagebackref,breaklinks,colorlinks,allcolors=iccvblue]{hyperref}
\usepackage{arydshln}
\usepackage{multirow}
\usepackage{adjustbox}
\usepackage{float}
\usepackage{placeins}
\usepackage{algorithm}
\usepackage{algpseudocode}
\def\paperID{1376} % *** Enter the Paper ID here
\def\confName{ICCV}
\def\confYear{2025}

\title{NeuSEditor: From Multi-View Images to Text-Guided Neural Surface Edits} 

\author{Nail Ibrahimli, Julian F. P. Kooij, Liangliang Nan\\
	Delft University of Technology\\
	Julianalaan 134, Delft 2628BL, The Netherlands\\
	{\tt\small \{n.ibrahimli, j.f.p.kooij, liangliang.nan\}@tudelft.nl}\\
}

\begin{document}
\twocolumn[{%
	\renewcommand\twocolumn[1][]{#1}%
	\maketitle
	\centering
	\includegraphics[width=0.925\linewidth]{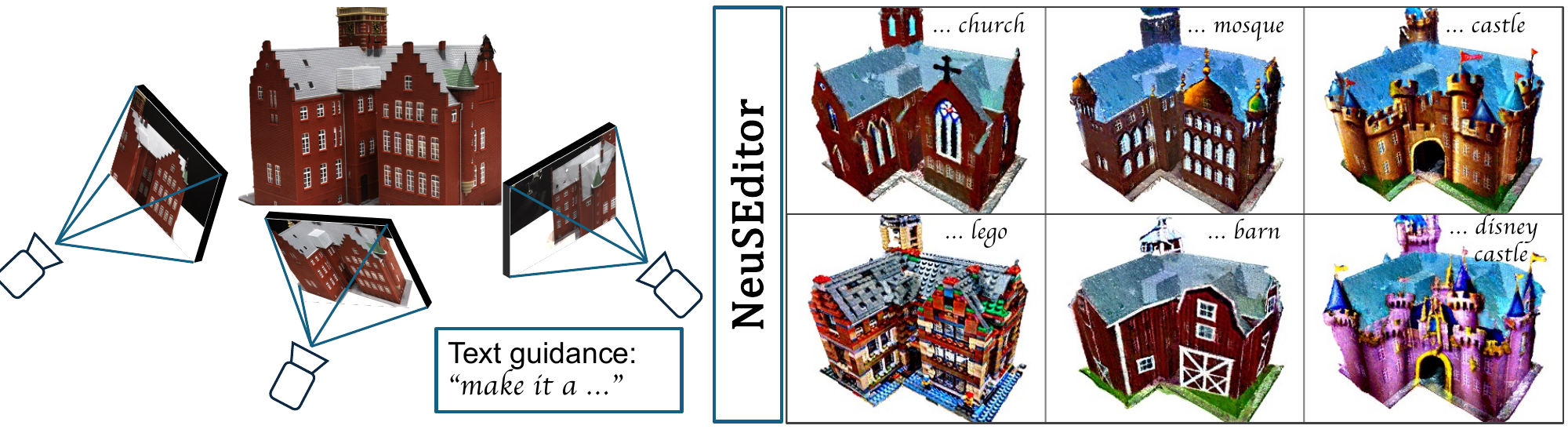}
	\captionof{figure}{ NeuSEditor is a novel method for text-guided neural surface editing from multi-view calibrated images. It generates an edited scene represented as a neural implicit field. Here, the sample inputs on the left are used to generate various 3D textured meshes. %with distinct structures guided by the text prompts.
		\vspace{2em}}
	\label{fig:teaser}
}]

\begin{abstract}
Implicit surface representations are valued for their compactness and continuity, but they pose significant challenges for editing. Despite recent advancements,  existing methods often fail to preserve identity and maintain geometric consistency during editing.  To address these challenges, we present NeuSEditor, a novel method for text-guided editing of neural implicit surfaces derived from multi-view images. NeuSEditor introduces an identity-preserving architecture that efficiently separates scenes into foreground and background, enabling precise modifications without altering the scene-specific elements.
Our geometry-aware distillation loss significantly enhances rendering and geometric quality.
Our method simplifies the editing workflow by eliminating the need for continuous dataset updates and source prompting.
NeuSEditor outperforms recent state-of-the-art methods like PDS and InstructNeRF2NeRF, delivering superior quantitative and qualitative results. 
For more visual results, visit: \href{https://neuseditor.github.io}{neuseditor.github.io}.
\end{abstract}  
\section{Introduction}
\label{sec:intro}
\begin{figure*}
    \centering
     \includegraphics[width=\textwidth]{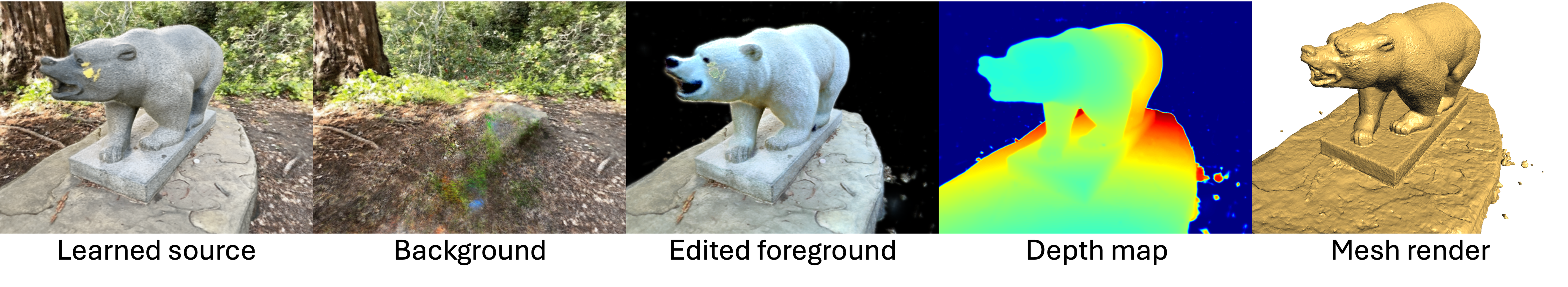}
    \caption{Our network architecture retains information about the identity (learned source) of the original scene, including background and foreground details, while simultaneously learning the edited (target) render and geometry. Text prompt: \textit{turn it to a polar bear}.}
    \label{fig:network_capability}
\end{figure*}
Recent advancements in neural rendering techniques have significantly enabled realistic 3D scene capture from multi-view images \cite{mildenhall2020NeRF,yariv2020multiview,Niemeyer2020CVPR,kerbl3Dgaussians}. 
These approaches are gaining popularity because of their ease of use and superior output quality over traditional photogrammetry \cite{nerfstudio,yu2022SDFStudio,li2023nerfacc,mueller2022instant,li2023neuralangelo,instant-nsr-pl}.
These methods, typically reliant on calibrated images obtained through Structure from Motion (SfM) pipelines, offer a more photorealistic digital twin of scenes \cite{schoenberger2016sfm}.

In parallel, diffusion-based generative models \cite{song2019NCSN,song2021SGM,ho2020DDPM,song2021DDIM,ho2021CFG} have sparked rapid progress in text-conditioned generation and editing across various domains, spanning 2D images~\cite{meng2022SDEdit,ramesh2022DALLE2,rombach2022StableDiffusion,ruiz2023DreamBooth,Brooks:2023InstructPix2Pix,alaluf2024cross}, 3D objects~\cite{nichol2022Point-E,jun2023ShapE,objaverse,objaverseXL,chou2022gensdf,chou2022diffusionsdf,shim2023diffusion}, and audio~\cite{ghosal2023Tango,yang2023DiffSound,huang2023Make-an-Audio}. 
Driven by large datasets of images and textual descriptions, 2D diffusion models have demonstrated strong generative capabilities~\cite{schuhmann2022laion,schuhmann2021laion,radford2021learning,li2022blip}. As a result, leveraging 2D diffusion priors has become a common approach for generating and editing 3D content from text~\cite{poole2023DreamFusion,lin2023Magic3D,chen2023Fantasia3D,zhu2023HiFA,Haque:2023InstructNeRF,raj2023dreambooth3d,wang2023prolificdreamer,metzer2023LatentNeRF,Koo:2024PDS,shi2023MVDream,liu2023zero,threestudio2023}.

However, editing implicit 3D scenes using text prompts presents several unresolved challenges. First, implicit geometry representations encapsulate the entire scene within neural network weights, often leading to inconsistent outputs that complicate precise local edits~\cite{Erkoc_2023_ICCV, lin2023Magic3D, Koo:2024PDS, zhuang2023dreameditor}. Second, view-based optimizations distort the scene from certain perspectives, introducing artifacts like floaters and multi-view inconsistencies ~\cite{philip2023floaters, shi2023MVDream, kant2024spad}. Furthermore, these issues, combined with the iterative nature of current methods, can result in catastrophic forgetting, where new edits overwrite or degrade previously learned scene features. 

In this paper, we propose a novel approach to address these challenges, thereby ensuring stable performance across diverse datasets.
%Unlike previous approaches that rely on additional loss terms \cite{Koo:2024PDS,hertz2023DDS} or dataset updates \cite{Haque:2023InstructNeRF,igs2gs}, our network architecture directly tackles these challenges by preserving scene identity and applying ``additive learning'' techniques \cite{hu2022lora,zhang2023adding} for edits.
%In this context, identity preservation refers to maintaining the essential and recognizable characteristics of the input scene throughout the editing process.
Unlike prior methods that depend on additional loss terms~\cite{Koo:2024PDS,hertz2023DDS} or require dataset updates~\cite{Haque:2023InstructNeRF,igs2gs}, our network architecture explicitly tackles the challenge of identity preservation (by maintaining the essential and recognizable characteristics of the input scene), while also incorporating ``additive learning'' techniques~\cite{hu2022lora,zhang2023adding} for incremental edits.
Furthermore, we introduce geometry-aware distillation which is effective with a single edit prompt~\cite{huberman2023FriendlyInversion,wu2023CycleDiffusion}. 
As shown in \cref{fig:teaser}, our proposed NeuSEditor enables Signed Distance Function (SDF)-based editing by utilizing multi-view images alongside simple editing text prompts. 
Our method is capable of generating accurate 3D textured meshes and renders across a variety of datasets, including DTU~\cite{aanaes2016large}, Blender~\cite{mildenhall2020NeRF}, and IN2N-data~\cite{Haque:2023InstructNeRF}, which experimental results have validated.

%In this paper, we introduce a novel approach that addresses these challenges while ensuring stable performance across diverse datasets. Unlike prior methods that depend on additional loss terms~\cite{Koo:2024PDS,hertz2023DDS} or require dataset updates~\cite{Haque:2023InstructNeRF,igs2gs}, our network architecture explicitly tackles the challenge of identity preservation by maintaining the essential and recognizable characteristics of the input scene, while also incorporating ``additive learning'' techniques~\cite{hu2022lora,zhang2023adding} for incremental edits. As shown in \cref{fig:teaser}, our proposed NeuSEditor enables Signed Distance Function (SDF)-based editing by leveraging multi-view images alongside simple text prompts. Consequently, our method generates accurate 3D textured meshes and high-quality renders, as validated on various datasets including DTU~\cite{aanaes2016large}, Blender~\cite{mildenhall2020NeRF}, and IN2N-data~\cite{Haque:2023InstructNeRF}. Moreover, we propose a geometry-aware distillation loss that proves effective even with a single edit prompt~\cite{huberman2023FriendlyInversion,wu2023CycleDiffusion}.

%The main contributions of this work are (1) a new implicit network architecture that preserves the identity of the input scene; (2) a novel geometry-aware distillation loss for 3D editing; and (3) a method with disentangled control for background and foreground editing.

The main contributions of this work are (1) a new implicit network architecture that preserves the identity of the input scene; (2) a novel geometry-aware distillation loss for 3D editing; and (3) a method that targets foreground edits without introducing undesired changes to the background, offering precise control over the editing process.

\section{Related works}
\label{sec:formatting}

{\bf{Differentiable rendering}} techniques have significantly advanced novel view synthesis and 3D reconstruction, starting with Neural Radiance Fields (NeRF). 
NeRF represents scenes as radiance fields for photorealistic rendering. However, NeRF is computationally intensive, requires long training times, and is unsuitable for precise geometry extraction due to its focus on volumetric color density rather than explicit surface modeling \cite{Niemeyer2020CVPR,Lombardi2019TOG,yariv2020multiview,mildenhall2020NeRF}.
To address the limitations in geometry representation, SDF-based volume rendering methods have been proposed \cite{yariv2020multiview,yariv2021volume,wang2021neus}, allowing for more accurate surface reconstruction. 

To accelerate rendering and training, explicit representations such as voxel-based approaches and 3D point or Gaussian-based methods have been introduced \cite{SunSC22,yu2021plenoctrees, xu2022point,kerbl3Dgaussians,ye2024gsplatopensourcelibrarygaussian}. Despite their performance advantages, these techniques require substantially more memory than implicit neural representations.
To balance quality and computational efficiency, methods like Instant Neural Graphics Primitives (Instant NGP) \cite{mueller2022instant} and Tensorial Radiance Fields
(TensorRF) \cite{Chen2022ECCV} have been introduced. 
Instant NGP accelerates the training of implicit neural representations using multiresolution hash encoding. 
TensorRF represents scenes with compact tensors, enabling efficient rendering and high-quality visual results. These approaches demonstrate that implicit representations can be fast and memory-efficient while maintaining high rendering quality.

%Our approach combines hash encodings from Instant NGP with NeuS's \cite{wang2021neus} volume rendering equations to achieve high-quality geometry and rendering efficiently. 
%While explicit representations simplify editing due to independent parameters, they require significant memory. 
%Implicit representations are compact and offer continuous detail but pose editing challenges due to entangled parameters and limited interpretability. 
%Our goal is to retain the advantages of implicit representations while enabling effective editing capabilities, overcoming these challenges without the storage overhead of explicit models.
%For example, in \cref{fig:teaser}, learning the identity of the scene requires approximately $28M$ parameters ($14M$ for the foreground and $14M$ for the background), with $14M$ additional parameters for each new edit.
% Thus, making \(N\) edits to the scene, we need to learn \(28M + 14M \times N\) parameters.
We aim to retain the benefits of implicit representations while enabling effective editing without the storage overhead of explicit models.  
For example, in \cref{fig:teaser}, learning the scene’s identity requires around $28M$ parameters ($14M$ for the foreground and $14M$ for the background), with an additional $14M$ per edit. Thus, making $N$ edits requires learning $28M + 14M \times N$ parameters.

{\bf{Text-guided neural rendering editing}} 
 gained popularity with the Instruct-Nerf2Nerf method \cite{Haque:2023InstructNeRF}. 
This method relies on the convergence of iterative dataset updates using the instruct-pix2pix model \cite{Brooks:2023InstructPix2Pix}. 
%Instruct-GS2GS \cite{igs2gs} is a follow-up work that achieves similar performance by replacing the underlying NeRF representation with Gaussian splatting.
Instruct-GS2GS \cite{igs2gs}  achieves similar performance by replacing the underlying NeRF representation with Gaussian splatting.

%DreamFusion introduced Score Distillation Sampling for text-to-NeRF reconstruction \cite{poole2023DreamFusion}. 
%Its goal is to optimize NeRF parameters so that the diffusion-based noise prediction on noisy rendered images matches the sampled noise. 
%The DDS work aims to extend the SDS loss for editing tasks, positing that the diffusion-based noise prediction on a noisy rendered source image should align with the noise prediction of a noisy target image \cite{hertz2023DDS}. 
%Recent studies have demonstrated the potential of utilizing stochastic latents for image editing, assuming that the stochastic latent structure of a noisy source image should be similar to that of the noisy target. \cite{wu2023CycleDiffusion,huberman2023FriendlyInversion}. 
%PDS incorporates this motivation as a loss term for NeRF editing \cite{Koo:2024PDS}. 

%DreamFusion \cite{poole2023DreamFusion} introduced Score Distillation Sampling (SDS), optimizing NeRF so that adding noise to its render and denoising with a diffusion model recovers the sampled noise.  
DreamFusion's text-to-NeRF method \cite{poole2023DreamFusion} introduced Score Distillation Sampling (SDS). This technique optimizes a NeRF by ensuring that when noise is added to its rendered output and then removed using a diffusion model, the original noise is accurately recovered.
DDS \cite{hertz2023DDS} extends SDS for editing, aligning noise predictions between noisy source and target images.  
%Recent works leverage stochastic posterior latents for editing, assuming structural similarity between noisy source and target images \cite{wu2023CycleDiffusion,huberman2023FriendlyInversion}.  
%PDS \cite{Koo:2024PDS} incorporates this idea as a loss term for NeRF editing.
Recent inversion methods leverage stochastic posterior latents for editing, assuming structural similarity between the noisy source and target posterior latents \cite{wu2023CycleDiffusion,huberman2023FriendlyInversion}. PDS \cite{Koo:2024PDS} integrates this observation as a loss term for NeRF editing.

Our method, unlike Instruct-Nerf2Nerf and Instruct-GS2GS, avoids iterative dataset updates. 
Unlike ours, both DDS and PDS require two text prompts to depict the source and target for noise and latent estimation. 
As shown in \cref{fig:network_capability}, unlike previous methods, our architecture explicitly disentangles the background from the foreground and separates identity (learned source) from edits (target). 
By retaining the scene identity within the network, our method not only alleviates catastrophic forgetting but also reduces the need for source prompting.

\section{Preliminaries}
In this section, we review the key concepts and notations that underpin our approach. In particular, we outline neural implicit representations and signed distance functions (SDFs), and discuss the diffusion-based editing methods.

\subsection{Neural implicit surfaces and rendering}
Neural implicit representations commonly encode scene structure and appearance in neural network weights instead of explicit geometry \cite{nerfstudio,yu2022SDFStudio}.  
For better capturing implicit geometry, the \emph{signed distance function} (SDF) serves as a key representation.  
 For any point $\mathbf{x} \in \mathbb{R}^3$, the SDF, denoted by $\delta(\mathbf{x})$, returns the signed distance to the nearest surface:
\begin{equation}
    \delta(\mathbf{x}) = \min_{\mathbf{y} \in \mathcal{S}} \|\mathbf{x} - \mathbf{y}\| - r,
\end{equation}
where $\mathcal{S}$ is the set of surface points and $r$ is a constant offset. The implicit surface is defined by the zero-level set $\{\mathbf{x} \mid \delta(\mathbf{x}) = 0\}$. This formulation allows for smooth and detailed surface extraction via \emph{volume rendering} \cite{mildenhall2020NeRF}. In particular, the color $\mathbf{c}$ along a ray $\mathbf{r}(t) = \mathbf{o} + t\mathbf{d}$ (with origin $\mathbf{o}$ and direction $\mathbf{d}$) is given by:
\begin{equation}
    \mathbf{c}(\mathbf{r}) = \int_{t_{near}}^{t_{far}} T(t) \, \sigma(\mathbf{r}(t)) \, \mathbf{c}(\mathbf{r}(t), \mathbf{d}) \, dt,
\end{equation}
with the transmittance defined as %$T(t) = \exp\left(-\int_{t_{near}}^{t} \sigma(\mathbf{r}(s)) \, ds\right),$
\begin{equation}
   T(t) = \exp\left(-\int_{t_{near}}^{t} \sigma(\mathbf{r}(s)) \, ds\right),
\end{equation}
where $\sigma(\mathbf{r}(t))$ is the volume density (which can be derived from $\delta$ \cite{yariv2021volume,wang2021neus}) and $\mathbf{c}(\mathbf{r}(t), \mathbf{d})$ is the view-dependent color.

Optimizing these representations can be computationally intensive. To address this, \emph{hash encoding} \cite{mueller2022instant,li2023neuralangelo} is employed to map 3D coordinates to a multiresolution hash grid storing feature vectors at grid vertices. These features are efficiently interpolated for any query point $\mathbf{x}$ and then fed into a lightweight MLP to predict both $\delta(\mathbf{x})$ and $\mathbf{c}(\mathbf{x}, \mathbf{d})$, significantly accelerating training and inference.

For \emph{foreground geometry}, our approach is inspired by Neuralangelo \cite{li2023neuralangelo}: we model the SDF using progressive hash encoding combined with NeuS's rendering equations. In contrast, for \emph{background regions} where SDF cannot be computed, we adopt an Inverted Sphere Parameterization \cite{zhang2020nerf++} along with hash encoding to effectively capture density fields in unbounded scenes.

\subsection{Diffusion models for 3D editing}
Diffusion models have recently achieved impressive results in text-to-image generation \cite{ho2020DDPM,rombach2022StableDiffusion}. In these models, an initial image ${x}_0$ is progressively perturbed by Gaussian noise over $t$ timesteps, yielding a noisy image ${x}_t$. The forward process is defined as:
\begin{equation}
    {x}_t = \sqrt{\bar{\alpha}_t}\,{x}_0 + \sqrt{1-\bar{\alpha}_t}\,\boldsymbol{\epsilon}, \quad \boldsymbol{\epsilon} \sim \mathcal{N}(0,\mathbf{I}),
\end{equation}
where $\bar{\alpha}_t = \prod_{s=1}^{t} \alpha_s$ and $\boldsymbol{\epsilon}$ is standard Gaussian noise.

The reverse process, which recovers ${x}_0$ from ${x}_t$, is performed via iterative denoising. In DDPM, a single reverse step is given by:
\begin{equation}
    \mu_{\phi}({x}_t, y) = \frac{1}{\sqrt{\alpha_t}} \left( {x}_t - \frac{1-\alpha_t}{\sqrt{1-\bar{\alpha}_t}}\,\epsilon_\phi({x}_t, y) \right), \label{eq:posterior_mean}
\end{equation}
\begin{equation}
    {x}_{t-1} = \mu_{\phi}({x}_t, y) + \sigma_t\,\mathbf{z}, \quad \mathbf{z} \sim \mathcal{N}(0,\mathbf{I}), \label{eq:x_t-1}
\end{equation}
where $\phi$ denotes the parameters of the diffusion network, $\epsilon_\phi({x}_t, y)$ is the predicted noise,  \( \sigma_t = \frac{1 - \bar{\alpha}_{t-1}}{1 - \bar{\alpha}_t} (1-\alpha_t)\), $y$ is text prompt, and \( \mathbf{z}\) represents standard Gaussian noise. 

{\bf{Score distillation sampling (SDS).}}
Building on diffusion models, \emph{DreamFusion} \cite{poole2023DreamFusion} introduced \textit{score distillation sampling} (SDS) to guide 3D optimization using 2D diffusion priors. Given a rendered image ${x}_0 = g(\theta)$ from 3D parameters $\theta$, SDS minimizes:
\begin{equation}
    \nabla_\theta \mathcal{L}_{\text{SDS}} = \mathbb{E}_{t,\epsilon} \Big[ w(t)\left(\epsilon_\phi({x}_t, y) - \epsilon\right) \frac{\partial {x}_0}{\partial \theta} \Big],
\end{equation}
where ${x}_t$ is a noised version of ${x}_0$, $y$ is a text prompt, and $w(t)$ is a time-dependent weighting function. While SDS enables text-to-3D generation, it struggles with \emph{editing tasks}, often causing catastrophic forgetting of the original scene \cite{poole2023DreamFusion, hertz2023DDS}.

\textbf{Delta denoising score (DDS).} 
To address SDS's limitations in editing, \cite{hertz2023DDS} proposed \textit{delta denoising score} (DDS). Instead of comparing $\epsilon_\phi({x}_t, y)$ to random noise $\epsilon$, DDS computes the \emph{difference} between noise predictions for source (${x}_0^{\text{src}}$) and target (${x}_0^{\text{tgt}}$) images:
\begin{equation}
\begin{split}
    \nabla_\theta \mathcal{L}_{\text{DDS}} &= \\\mathbb{E}_{t,\epsilon_t} &\Big[ w(t)\left(\epsilon_\phi({x}_t^{\text{tgt}}, y^{\text{tgt}}) - \epsilon_\phi({x}_t^{\text{src}}, y^{\text{src}})\right) 
    \frac{\partial {x}_0^{\text{tgt}}}{\partial \theta} \Big].
\end{split}
\end{equation}
While DDS demonstrated potential in 2D image editing, its direct application to 3D scene editing often leads to failure to preserve structural coherence \cite{Koo:2024PDS}.

\textbf{Posterior distillation sampling (PDS).}
The stochastic latent \(\mathbf{z}\) in \cref{eq:x_t-1}, often termed the \emph{posterior stochastic latent}, encodes structural information critical for preserving scene coherence during editing \cite{huberman2023FriendlyInversion, wu2023CycleDiffusion}. Rearranging \cref{eq:x_t-1}, \(\mathbf{z}\) is derived as:  
\begin{equation}
    \mathbf{z} = \frac{{x}_{t-1} - \mu_{\phi}({x}_t, y)}{\sigma_t}. \label{eq:stochastic_latent}
\end{equation}

Building on this insight, \textit{posterior distillation sampling} (PDS) \cite{Koo:2024PDS} adapts stochastic latent alignment to 3D neural editing by minimizing the discrepancy between source and target scene posterior latents:  
\begin{equation}
    \mathcal{L}_{\text{PDS}} = \mathbb{E}_{t, \epsilon_t, \epsilon_{t-1} \sim \mathcal{N}(0,\mathbf{I})} \left\| \mathbf{z}_t^{\text{tgt}} - \mathbf{z}_t^{\text{src}} \right\|^2_2, 
\end{equation}
where \(\mathbf{z}_t^{\text{src}}\) and \(\mathbf{z}_t^{\text{tgt}}\) denote the posterior latents of the source (unedited) and target (edited) scenes.

%%%%%%%%%%%%%%%%%%%%%%%METHOD%%%%%%%%%%%%%%%%%%%%%%%%%%%%%
\section{Method}
\label{sec:method}
\begin{figure*}[!h]
    \centering
    \includegraphics[width=0.94\textwidth]{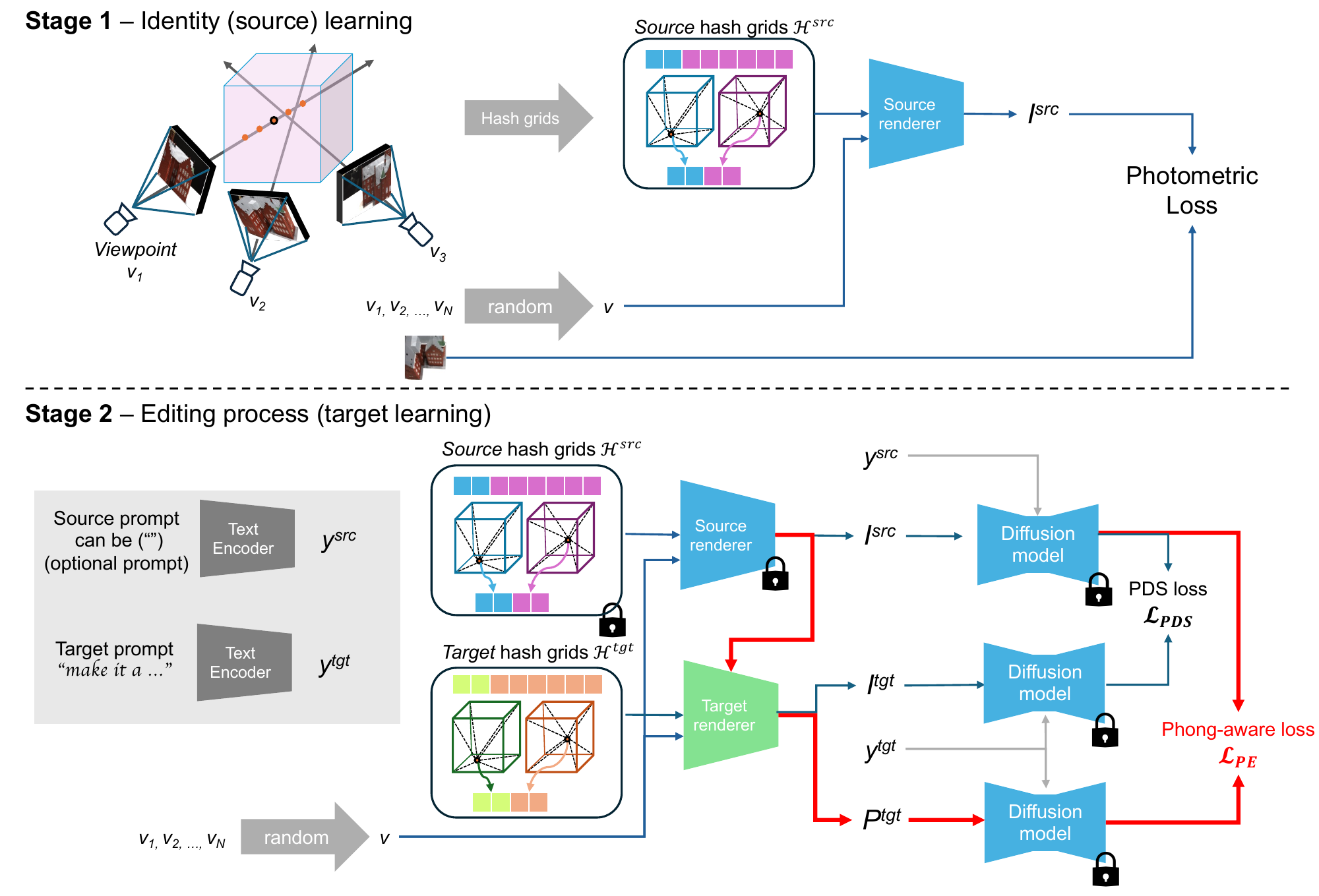}
    %\caption{Overview of our two-stage scene editing pipeline. In the first stage (source learning), the network learns the identity of the scene, capturing its geometry and appearance using (SDF-based) volume rendering with progressive hash encodings. In the second stage (target learning), an editing process modifies the scene. A target renderer, conditioned on the source renderer, enables effective scene edits, guided by geometry-aware distillation loss. Here, $\mathcal{I}^{\text{src}}$ and $\mathcal{I}^{\text{tgt}}$ denote the source and target renderings, respectively, while $\mathcal{P}^{\text{tgt}}$ represents Phong shading. Within the pipeline flow, the red highlights clearly emphasize the transfer of knowledge between the source and target, the integration of Phong shading, and its role in the loss formulation.}
    \caption{Overview of our two stage scene editing pipeline. In the first stage (source learning) the network captures the scene's identity, including its geometry and appearance, using SDF based volume rendering with progressive hash encodings. In the second stage (target learning) a target renderer, conditioned on the source renderer and guided by a geometry aware distillation loss, edits the scene. Here, $I_{\mathrm{src}}$ and $I_{\mathrm{tgt}}$ denote the source and target renderings respectively, while $P_{\mathrm{tgt}}$ represents Phong shading. Within the pipeline, the red arrows highlights the transfer of knowledge between source and target, the integration of Phong shading, and its role in the loss formulation.}

    \label{fig:pipeline}
\end{figure*}
%In this section, we detail our network architecture and our main optimization criterion and provide the rationale behind our design choices.
%\subsection{Network architecture}
\label{sec:net_arch}
%Our method adopts a similar approach to neural rendering techniques~\cite{mildenhall2020NeRF,yariv2021volume} by querying 3D points along with the view direction as input, producing a colored implicit field value. \cref{fig:pipeline} illustrates an overview of our pipeline. 
Our method follows a two-stage approach. In the first stage, our network learns the identity of the source scene depicted in the input images. Similar to neural rendering techniques~\cite{mildenhall2020NeRF,yariv2021volume}, the network queries 3D points along with their corresponding view directions to generate a colored implicit field. In the second stage, after capturing the scene identity, the network shifts its focus to the editing process (target learning) while preserving the original identity. \cref{fig:pipeline} provides an overview of our pipeline.
\subsection{Identity (source) learning}
%To capture the identity of the scene, we employ progressive hash encodings to obtain interpolated 3D hash grid features \cite{li2023neuralangelo}. Initially, these features are fed into an MLP-based geometry module. The output of this module (including the SDF, the gradient of the SDF, and a feature vector) is then combined with the input view direction to feed into the rendering network. We leverage photometric loss and eikonal loss \cite{yaron_igr_icml2020} to effectively learn the ``identity'', which encapsulates the parameters of the source hash grids as well as the weights of the source geometry and rendering networks.
Our approach captures the identity of a scene, its geometry and appearance, by combining progressive hash encodings \cite{li2023neuralangelo} with SDF-based volume rendering \cite{wang2021neus}. At the core of this method is an MLP network that predicts both the signed distance (SDF) and color values for each 3D point. We train the MLP-based source renderer using photometric and eikonal losses \cite{yaron_igr_icml2020}. Through this process, the scene's identity is fully encoded in the parameters of the hash grids and the weights of the MLP-based source renderer networks.

\subsection{Editing process (target learning)}
%We utilize an MLP network to learn the target geometry.
%We employ a second set of hash grids specifically for editing tasks, which allows us to obtain interpolated 3D hash grid features.
%Initially, we input interpolated target hash grid features, and at the hidden layers, we further condition the learned representations with source geometry features. Additionally, we incorporate a residual connection for the target hash features at the hidden layers. This design encourages the network to effectively fuse source geometric features with the target hidden features at each grid point.  

%This fusion submodule produces the geometry features for the edit, including the SDF, the gradient of the SDF, and a feature vector. The output of this fusion module is then passed to the rendering module along with the color of the identity. To enhance the editing process, we utilize an ``additive learning'' technique  \cite{hu2022lora,zhang2023adding} by initializing both the fusion and editing render submodules so that, at the start of the editing process, the target geometry and color correspond to the identity. This initialization strategy encourages the network to learn as necessary during the editing process.

To edit the geometry, we introduce a second set of 3D hash features. These features are fed into an MLP-based target renderer that is conditioned on the source renderer, which encodes SDF geometry and geometric feature vector at each 3D point. This conditioning allows the target renderer to access richer features for effective scene editing.

We employ an additive learning strategy~\cite{hu2022lora,zhang2023adding} to enhance the editing process. Specifically, we initialize the target renderer networks so that at the start of the editing process, the target geometry and color match the identity. This initialization encourages target networks to focus on learning the essential modifications during the editing process. The target renderer networks are trained using a geometry-aware distillation process that leverages a diffusion model.

\subsection{Geometry-aware distillation}
\newcommand{\B}{\mathbf}
%The single-step forward process of diffusion can be expressed using the following equation \cite{ho2020DDPM}:
%\begin{equation}
%x_t = \sqrt{\bar{\alpha}_t} x_0 + \sqrt{1 - \bar{\alpha}_t} \epsilon ,
%\end{equation}
%where \( x_t \) is the noisy image at timestep \( t \); \( x_0 \) is the original image; \( \bar{\alpha}_t = \prod_{s=1}^{t} \alpha_s \) is the cumulative variance, and \( \epsilon \sim \mathcal{N}(0, I) \) represents Gaussian noise. 
%The single-step backward process in DDPM during the generation phase can be described as follows:
%\begin{equation}
%      \mu_{\phi}(x_t, t) = \frac{1}{\sqrt{\alpha_t}} \left( x_t - \frac{1 - \alpha_t}{\sqrt{1 - \bar{\alpha}_t}} \epsilon_\phi(x_t, t) \right),\label{eq:posterior_mean} 
%\end{equation}
%\begin{equation}
 %x_{t-1} = \mu_{\phi}(x_t, t) + \sigma_t \mathbf{z} ,
 %\label{eq:x_t-1}
 %\end{equation}
%where \( \mu_{\phi}(x_t, t) \) is the posterior mean; \( \alpha_t \) is the variance schedule; \( \bar{\alpha}_t = \prod_{s=1}^{t} \alpha_s \) is the cumulative variance; \( \epsilon_\phi(x_t, t) \) is the predicted noise; \( \sigma_t = \frac{1 - \bar{\alpha}_{t-1}}{1 - \bar{\alpha}_t} (1-\alpha_t)\), and \( \mathbf{z} \sim \mathcal{N}(0, I) \) represents standard Gaussian noise. 
%\cref{eq:posterior_mean,eq:x_t-1} illustrate how the posterior mean is utilized to reconstruct \( x_{t-1} \) from the noisy observation \( x_t \). 

We present a novel, geometry-aware method for stochastic latent distillation. Our approach begins by computing the stochastic latent variables as defined in  \cref{eq:stochastic_latent}:
\begin{align}
    \B{z}_{t}^{\text{src}} = \B{z}_t({x}_0^{\text{src}}, y^{\text{src}}; &\epsilon_\phi), \quad y^{\text{src}} \in \{ \text{source prompt}, \emptyset \} \label{eq:source_prompt} \\
    \B{z}_{t}^{\text{tgt}} &= \B{z}_t({x}_0^{\text{tgt}}, y^{\text{tgt}}; \epsilon_\phi) \\
    \hat{\B{z}}_{t}^{\text{tgt}} &= \hat{\B{z}}_t(\text{Phong}_0^{\text{tgt}}, y^{\text{tgt}}; \epsilon_\phi),
\end{align}
where \(\B{z}_{t}^{\text{src}}\) represents the stochastic latent of the source image at timestep \(t\), with \(y^{\text{src}}\) denoting the source prompt. The variable \(y^{\text{tgt}}\) indicates the editing prompt, while \(\B{z}_{t}^{\text{tgt}}\) corresponds to the latent representation for the target image. 
Additionally, to enhance and promote geometric edits, we employ a Phong shading renderer. \(\hat{\B{z}}_{t}^{\text{tgt}}\) captures the stochastic latent associated with Phong shading of the target geometry. Computing Phong shading requires target SDF gradients. We find \textit{numerical} gradients work better than \textit{analytical} ones and provide details in the supplementary material.
Since the Diffusion model requires a three-channel image, we convert the Phong shading to a three-channel image by simply repeating it across all three channels. 

Other methods~\cite{hertz2023DDS,Nam_2024_CDS,Koo:2024PDS} typically involve two prompts to incorporate input (source) information. However, using two different prompts without explicit identity-preserving mechanism can easily steer the optimization in different directions. To alleviate this issue, PDS \cite{Koo:2024PDS} employs two very similar prompts with keyword replacement or addition for editing purposes. In contrast, our method does not even require a source prompt, as we address identity preservation directly through the network architecture.

Our overall objective can be expressed as follows:
\begin{equation}
   \mathcal{L}_{\text{PDS}} = \mathbb{E}_{t \sim U(0,1); \, \epsilon_t, \epsilon_{t-1} \sim N(0, I)} \left\| \mathbf{z}_{t}^{\text{tgt}} - \mathbf{z}_{t}^{\text{src}} \right\|_2^2   
\end{equation}
\begin{equation}
   \mathcal{L}_{\text{PE}} = \mathbb{E}_{t \sim U(0,1); \, \epsilon_t, \epsilon_{t-1} \sim N(0, I)} \left\| \hat{\mathbf{z}}_{t}^{\text{tgt}} - \mathbf{z}_{t}^{\text{src}} \right\|_2^2   
\end{equation}
\begin{equation}
   \mathcal{L}_{\text{PEPDS}} = \lambda_{\text{PDS}} \mathcal{L}_{\text{PDS}} + \lambda_{\text{PE}} \mathcal{L}_{\text{PE}},
\end{equation}
where \(\mathcal{L}_{\text{PEPDS}}\) is the overall loss function, \(\lambda_{\text{PDS}}\) and \(\lambda_{\text{PE}}\) are weighting factors for the first and second terms, respectively. \(\mathcal{L}_{\text{PDS}}\) matches the stochastic latents of the source and target images, while \(\mathcal{L}_{\text{PE}}\) aligns the stochastic latents of the source and target Phong shading images. The addition of Phong shading-based distillation significantly reduces floaters in geometry (see also \cref{sec:ablation}). 
To avoid the costly computation associated with the Diffusion UNet~\cite{rombach2022StableDiffusion,runwayml-stable-diffusion}, we directly compute the gradient of the loss function  with respect to the (geometry and) renderer parameters $\theta$, as suggested in SDS~\cite{poole2023DreamFusion}, while omitting the U-Net Jacobian:
\begin{equation}
  \nabla_{\theta}\mathcal{L}_{\text{PDS}} =  \mathbb{E}_{t, \epsilon_t, \epsilon_{t-1}} \left[ \mathit{w}(t)  ( {\mathbf{z}}_{t}^{\text{tgt}} - \mathbf{z}_{t}^{\text{src}} )\frac{\partial x_{0}^{\text{tgt}}}{\partial\theta}\right]
\end{equation}
\begin{equation}
  \nabla_{\theta}\mathcal{L}_{\text{PE}} =  \mathbb{E}_{t, \epsilon_t, \epsilon_{t-1}} \left[  \mathit{w}(t)  ( \hat{\mathbf{z}}_{t}^{\text{tgt}} - \mathbf{z}_{t}^{\text{src}} )\frac{\partial x_{0}^{\text{tgt}}}{\partial\theta}\right]
\end{equation}
\begin{equation}
   \nabla_{\theta}\mathcal{L}_{\text{PEPDS}} = \lambda_{\text{PDS}} \cdot \nabla_{\theta}\mathcal{L}_{\text{PDS}} + \lambda_{\text{PE}} \cdot \nabla_{\theta}\mathcal{L}_{\text{PE}},
\end{equation}
where $\mathit{w(t)}$ is a weighting function depending on timestep \cite{ho2020DDPM} and $ x_{0}^{\text{tgt}}$ is the target image. 
\section{Experiments}

\subsection{Experimental Details}
\label{sec:hyperparam}
\begin{figure*}[h!]
    \centering
    \includegraphics[width=0.925\textwidth]{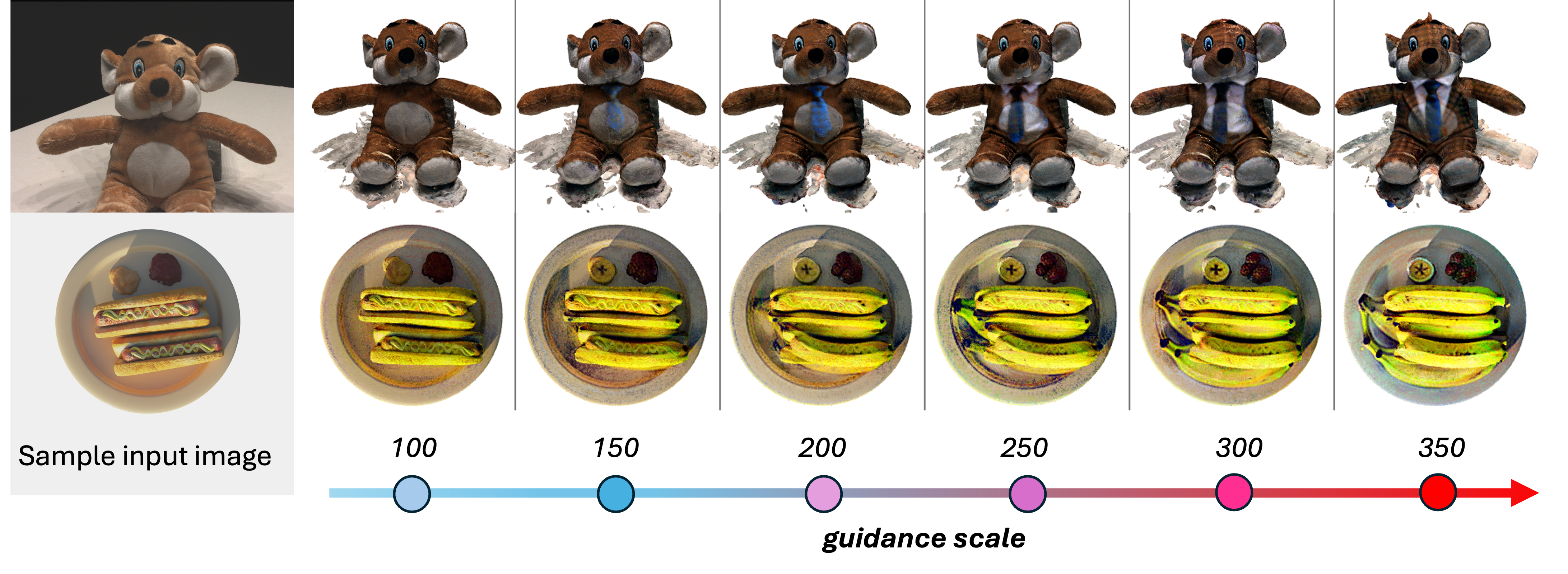}
    \caption{Influence of hyperparameter guidance scale. Left: two sample input images from the DTU dataset (scan105 scene) and the Blender dataset (hotdog scene), respectively. Right: the edited mesh renderings of the corresponding scenes at guidance scales from 100 to 350. The text prompt for the top row is `\textit{add suit}', while the prompt for the bottom row is `\textit{make it bananas}'.}
    \label{fig:guidance_scale}
\end{figure*}

Experiments ran on a single A100 GPU, with training times (1.5–6h) comparable to PDS\textsubscript{NeRF}. Duration depends on (1) image resolution, (2)  foreground pixel coverage and (3) prompt complexity (e.g., Bear: 1.5h; DTU: 4–6h, including reconstruction + editing). One of the crucial hyperparameters is the guidance scale~\cite{ho2021CFG}. As illustrated in ~\cref{fig:guidance_scale}, a guidance scale of 350 generally yields superior results. It is noteworthy to observe how increasing the guidance scale turns the tie to the suit (top row) and gives the ketchup and mustard a fruity appearance (bottom row). 
\subsection{Ablation study}
\label{sec:ablation}
\begin{figure}[t!]
    \centering
\includegraphics[width=0.915\columnwidth]{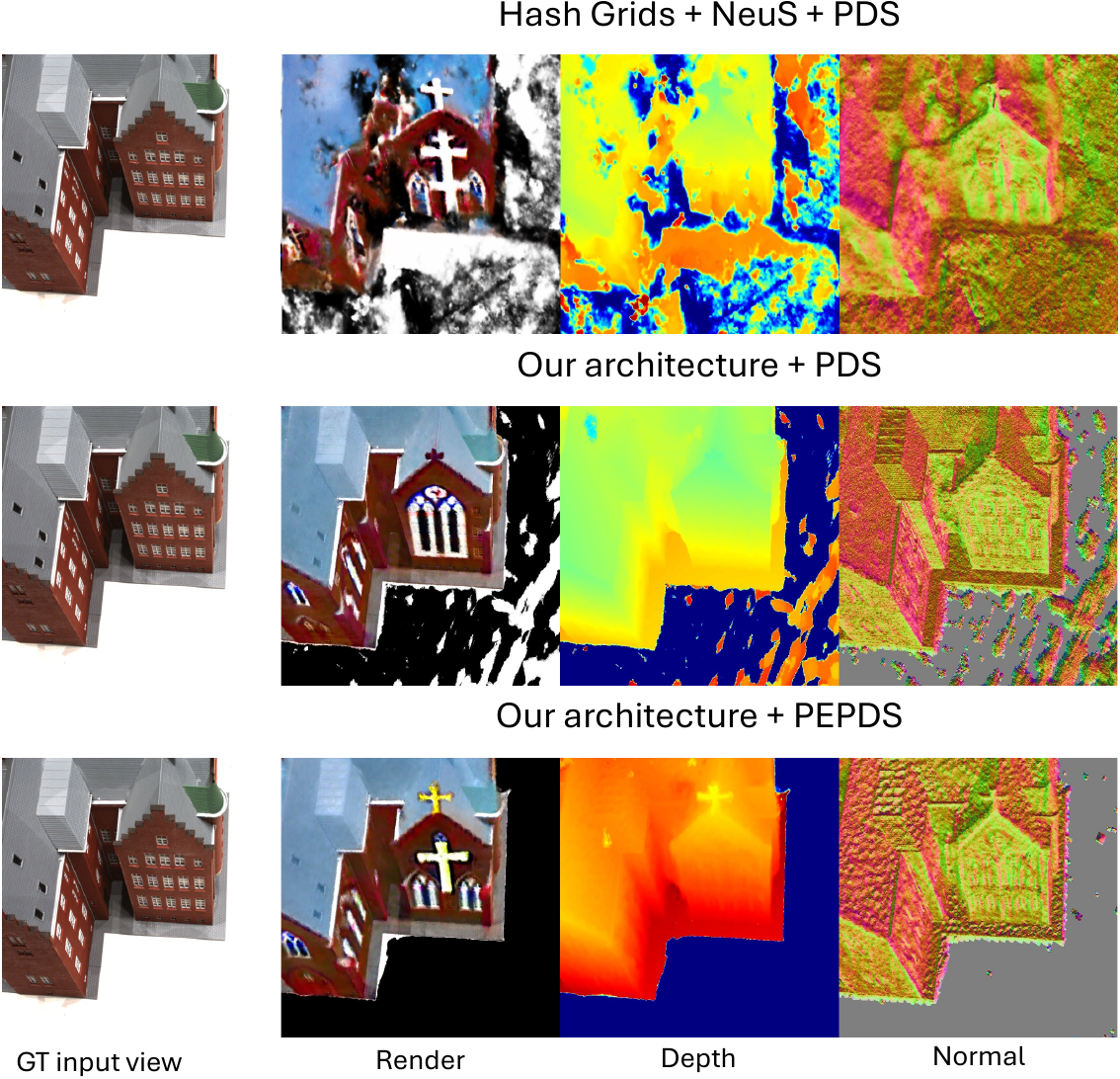}
    \caption{The left column displays an input view from the DTU dataset scan24, while the right column shows the foreground render, depth map, and normal map. The top row presents results from the naive Hash Grids + NeuS + PDS loss approach. The middle row shows our architecture with PDS loss, and the bottom row shows our architecture with Phong Enhanced PDS loss\\
    ($\lambda_\text{PDS}=1.$ and $\lambda_\text{PE} = 0.2$). The text prompt is `\textit{make it a church}'.}
    \label{fig:ablation}
\end{figure}
We conducted an ablation study to validate the contributions of our architecture and geometry-aware loss. In all experiments, we employed simple Hash Grids and analytical SDF gradient computations to exclusively assess the architecture and our loss. \cref{fig:ablation} summarizes the qualitative results.

\textbf{Baseline.} The top row shows the naive baseline, which uses Hash Grids for positional encoding~\cite{mueller2022instant}, NeuS as the renderer~\cite{wang2021neus}, and PDS as the distillation loss~\cite{Koo:2024PDS}. The render view, depth map, and normal map reveal that this combination fails to preserve the original scene's identity and produces noisy geometry.

\textbf{Architecture Enhancement.} The middle row demonstrates that incorporating our architecture significantly improves geometry and rendering quality, although some floating artifacts~\cite{philip2023floaters} remain.

\textbf{Loss Improvement.} The bottom row highlights the impact of our loss objective: by employing Phong-aware distillation, the fidelity of fine details is enhanced and the number of floaters is drastically reduced. 
\begin{table}[h]
    \centering
    \begin{tabular}{lccccc}
        \hline
        Metric & Backbone & Baseline & Arch + PDS & Ours \\
        \hline
        \multirow{2}{*}{clip $\uparrow$} & ViT/16  & 0.268 & \cellcolor{yellow!25}0.276 & \cellcolor{red!25}0.283 \\
                             & ViT/32  & 0.261 & \cellcolor{yellow!25}0.270 & \cellcolor{red!25}0.276 \\
        \hline
        \multirow{2}{*}{lpips $\downarrow$} & Alex    & 0.721 & \cellcolor{red!25}0.685 & \cellcolor{yellow!25}0.700 \\
                               & VGG     & 0.720 & \cellcolor{red!25}0.676 & \cellcolor{yellow!25}0.683 \\
        \hline
        \multirow{2}{*}{lpips (MV) $\downarrow$} & Alex    & 0.133 & \cellcolor{yellow!25}0.112 & \cellcolor{red!25}0.101 \\
                               & VGG     & 0.231 & \cellcolor{yellow!25}0.187 & \cellcolor{red!25}0.177 \\
        \hline
    \end{tabular}
% \caption{Quantitative ablation study results using CLIP and LPIPS. The last row evaluates Multi-view consistency between consecutive frames using LPIPS distance metric.}
\caption{Quantitative ablation study results based on CLIP and LPIPS metrics. The final row reports the evaluation of multi-view consistency between consecutive frames using the LPIPS  metric.}
\label{tab:ablation}
\end{table}

We evaluated the ablations using both CLIP \cite{radford2021learning} and LPIPS \cite{zhang2018perceptual} metrics (see \cref{tab:ablation}). Our analysis of foreground renderings shows that our architecture and loss function improve the CLIP similarity score. To evaluate catastrophic forgetting, we trained a separate NeRF model to render videos from input images. Then, we computed the LPIPS distance between the edited rendering and the reference rendering from the separate NeRF model (“input” scene). The results show a significant reduction in distance, indicating better identity preservation, which is due to our architecture. Additionally, we calculate the LPIPS distance between consecutive frames, expecting floaters and inconsistencies to increase the distance. Results show that both the proposed architecture and loss contribute to better consistency.

\subsection{Dataset and benchmark}
\label{sec:dataset}
Our evaluation spans three datasets: DTU~\cite{aanaes2016large} (6 scenes, 11 experiments), Blender~\cite{mildenhall2020NeRF} (3 scenes, 7 experiments), and IN2N~\cite{Haque:2023InstructNeRF} (2 scenes, 6 experiments), totaling 11 scenes and 34 editing experiments. 
To ensure fair evaluation, we rendered videos using spherical interpolation of camera poses, setting the radius to the average distance from the scene origin. This approach was applied to all DTU scenes and the \textit{bear} scene from the IN2N dataset. 
The \textit{person} scene in the IN2N dataset was not suitable for this strategy, so we adopted a path similar to the default IN2N.
For Blender, we followed the predefined test dataset trajectory.
\begin{table}[h]
    \centering
    \begin{tabular}{lcccccc}
        \hline
        Metric & Backbone & IN2N & PDS\textsubscript{NERF} & PDS\textsubscript{GS} & Ours \\
        \hline
        \multirow{2}{*}{clip $\uparrow$} & ViT/16 & 0.263 & 0.272 & \cellcolor{yellow!25}0.296 & \cellcolor{red!25}0.297 \\
                             & ViT/32 & 0.264 & 0.272 & \cellcolor{yellow!25}0.293 & \cellcolor{red!25}0.294 \\
        \hline
        \multirow{2}{*}{lpips $\downarrow$} & Alex   & 0.427 & \cellcolor{yellow!25}0.412 & 0.432 & \cellcolor{red!25}0.306 \\
                               & VGG    & 0.443 & \cellcolor{yellow!25}0.415 & 0.438 & \cellcolor{red!25}0.344 \\
        \hline
    \end{tabular}
\caption{The top row shows text-image similarity, and the bottom shows perceptual similarity between the input and edited scenes.}
    \label{tab:method_comparison}
\end{table}
\subsubsection{Objective Metrics}
\label{sec:objective_metrics}

To quantify editing performance, we computed text-image consistency and perceptual similarity-based metrics across all methods (\cref{tab:method_comparison}). Specifically, we used CLIP scores with PDS text prompts over 34 editting tasks. Our method performs on par with the explicit approach (PDS\textsubscript{Splat}) while outperforming implicit methods (IN2N, PDS\textsubscript{NeRF}).

Additionally, similar to our ablation experiments, we trained a separate NeRF model to render videos from input images across all experiments. For each of the 34 editting tasks, we computed the LPIPS distance between the 3D-edited rendering and the reference (unedited) rendering from the separate NeRF model. High LPIPS distances indicate catastrophic forgetting in the edited scene. Our analysis reveals that our method preserves the original scene significantly better than competing approaches.

\subsubsection{User study}
\label{sec:user_study}

We conducted a user study with 41 participants, each completing 34 experiments. In each experiment, participants viewed rendering videos from our model, PDS\textsubscript{NeRF}~\cite{Koo:2024PDS,nerfstudio}, PDS\textsubscript{Splat}~\cite{Koo:2024PDS,ye2024gsplatopensourcelibrarygaussian}, and IN2N~\cite{Haque:2023InstructNeRF,nerfstudio}, alongside original scene images and editing text prompts. The video order was randomized to eliminate bias, and participants ranked each video on a scale from 0 (lowest quality) to 3 (highest quality). The evaluation prompt stated: 

\textit{``Considering the original scene images and editing text prompts, assign an overall score to the following scenes in terms of geometry, editing quality, and clarity while preserving the features of the original scene.''}

\begin{figure*}[h!]
    \centering
\includegraphics[width=0.85\textwidth]{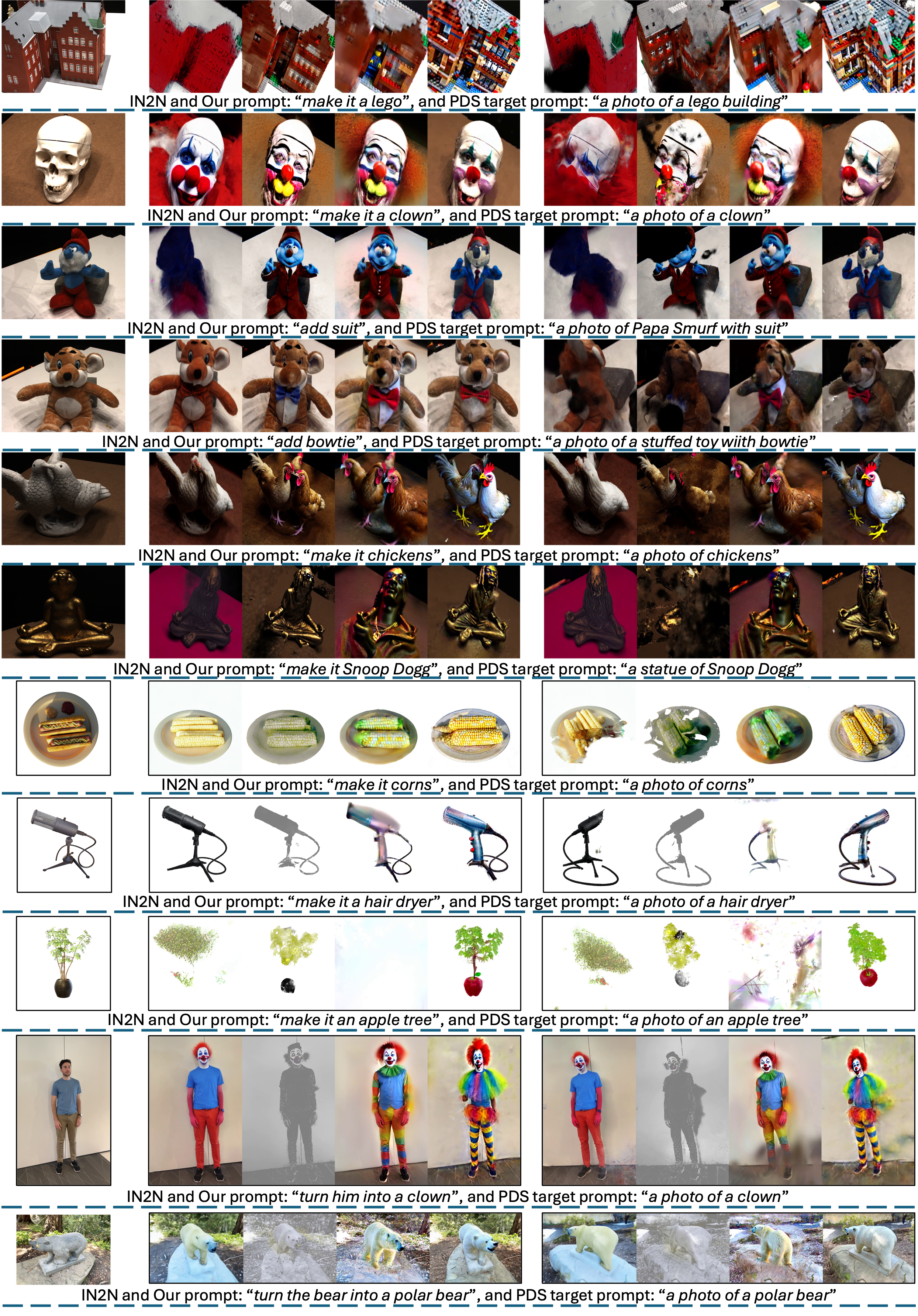}
    \caption{Comparison between IN2N, PDS\textsubscript{NeRF}, PDS\textsubscript{Splat}, and our method (in this order) across two frames. Edit prompts for ours and IN2N, and the target prompts for the PDS models are also provided below each row.
}
    \label{fig:survey}
\end{figure*}

\cref{tab:results,tab:average_results_per_scene} summarize the results across datasets. \cref{fig:survey} presents survey samples, showing a reference image and two frames from four models: one with minimal noise and another highlighting floating artifacts that may affect user choices.  
Based on both quantitative and qualitative results, our method and PDS\textsubscript{Splat} outperform other methods in DTU, whereas IN2N struggles with degenerate cases, and PDS\textsubscript{NeRF} exhibits a substantial amount of floating artifacts. 

In Blender, our method and PDS\textsubscript{Splat} again outperform on \textit{hotdog} and \textit{mic} scenes, while PDS\textsubscript{NeRF} fails. IN2N surpasses PDS\textsubscript{NeRF} due to fewer floating artifacts. In \textit{ficus}, PDS\textsubscript{Splat} encounters multi-view inconsistencies, IN2N struggles, and PDS\textsubscript{NeRF} aligns better with text prompts.

For the IN2N dataset, IN2N achieves the best performance. In \textit{person}, all methods produce clean results, with users favoring PDS\textsubscript{Splat} and IN2N for identity preservation. In \textit{bear}, our method and IN2N yield comparable results, with users slightly preferring ours for mitigating the ``Janus artifact.'' PDS\textsubscript{Splat} struggles with multi-view consistency, and PDS\textsubscript{NeRF} displays monotonic color artifacts. See supplementary and the \href{https://neuseditor.github.io/survey_experiments/survey_index.html}{survey page replica} for details.

\begin{table}[t]
    \centering
    \begin{tabular}{@{}lcccc@{}}
        \toprule
        Dataset & IN2N  & PDS\textsubscript{NeRF} & PDS\textsubscript{Splat} & Ours \\ 
        \midrule
        DTU & 0.52 & 1.07 & \cellcolor{yellow!25}1.87 & \cellcolor{red!25}2.54 \\
        Blender & 1.08 & 0.71 & \cellcolor{yellow!25}1.53 & \cellcolor{red!25}2.69 \\
        IN2N-data & \cellcolor{red!25}2.38 & 0.63 & 1.39 & \cellcolor{yellow!25}1.61 \\
        \hdashline
        Avg. per dataset & 1.33 & 0.80 & \cellcolor{yellow!25}1.60 &  \cellcolor{red!25}2.28 \\ 
        Avg. per experiment & 1.00 & 0.92 & \cellcolor{yellow!25}1.71 &       \cellcolor{red!25}2.41 \\ 
        \bottomrule
    \end{tabular}
    \caption{Survey results across datasets. The last rows show averages per dataset and all 34 experiments. (Higher is better $\uparrow$)}
    \label{tab:results}
\end{table}
\begin{table}[h t]
    \centering
    \begin{tabular}{@{}lcccc@{}}
        \toprule
        Dataset - Scene & IN2N & PDS\textsubscript{NeRF} & PDS\textsubscript{Splat} & Ours \\ \midrule
        DTU - Scan24 & 0.18 & 1.26 & \cellcolor{yellow!25}1.97 & \cellcolor{red!25}2.59 \\
        DTU - Scan65 & 0.92 & 0.80 & \cellcolor{yellow!25}1.58 & \cellcolor{red!25}2.69 \\
        DTU - Scan83 & 0.07 & \cellcolor{yellow!25}1.95 & 1.76 & \cellcolor{red!25}2.22 \\
        DTU - Scan105 & 0.55 & 1.23 & \cellcolor{yellow!25}1.80 & \cellcolor{red!25}2.41 \\
        DTU - Scan106 & 0.52 & 1.18 & \cellcolor{yellow!25}1.67 & \cellcolor{red!25}2.62 \\
        DTU - Scan110 & 0.83 & 0.30 & \cellcolor{yellow!25}2.41 & \cellcolor{red!25}2.46 \\
        Blender - Hotdog & 1.09 & 0.39 & \cellcolor{yellow!25}1.77 & \cellcolor{red!25}2.76 \\
        Blender - Mic & 1.28 & 0.17 & \cellcolor{yellow!25}2.07 & \cellcolor{red!25}2.48 \\
        Blender - Ficus & 0.93 & \cellcolor{yellow!25}1.28 & 1.00 & \cellcolor{red!25}2.78 \\
        IN2N - Person & \cellcolor{red!25}2.37 & 0.65 & \cellcolor{yellow!25}2.20 & 0.79 \\
        IN2N - Bear & \cellcolor{yellow!25}2.39 & 0.60 & 0.58 & 
        \cellcolor{red!25}2.43 \\
        \hdashline
        Avg. per scene & 1.01 & 0.89 & \cellcolor{yellow!25}1.71 & 
        \cellcolor{red!25}2.38 \\ \bottomrule
    \end{tabular}
    \caption{Average user score per scene. (Higher is better $\uparrow$)}    \label{tab:average_results_per_scene}
\end{table}
\subsection{Limitations}
\label{sec:user_limitations} NeuSEditor shares some common limitations with 3D generative AI pipelines. First, these pipelines generally require high CFG weights, which can steer optimization toward oversaturated colors \cite{poole2023DreamFusion}. Additionally, our method is constrained by the inherent biases of underlying diffusion models, making it unable to completely avoid Janus artifacts or content drift issues \cite{shi2023MVDream,kant2024spad}. 
\section{Conclusion}
In this paper, we presented NeuSEditor, a novel approach for text-guided editing of neural implicit surfaces from multi-view images. NeuSEditor introduces a novel identity-preserving architecture combined with a geometry-aware distillation loss, enabling high-quality 3D content editing that outperforms state-of-the-art methods in terms of rendering fidelity and geometric consistency. Our approach preserves scene identity while providing precise edit control. 
Our empirical results demonstrate that NeuSEditor effectively mitigates common issues such as catastrophic forgetting and degenerate outputs, such as those with monotonic colors, which are common pitfalls in other implicit surface-based methods. 
We believe that the novel architecture and Phong-based distillation could inspire new approaches for robust and compact 3D content editing, offering a foundation for future advancements in this area.
{
	\small
	\bibliographystyle{ieeenat_fullname}
	\bibliography{main}

\begin{thebibliography}{72}
\providecommand{\natexlab}[1]{#1}
\providecommand{\url}[1]{\texttt{#1}}
\expandafter\ifx\csname urlstyle\endcsname\relax
  \providecommand{\doi}[1]{doi: #1}\else
  \providecommand{\doi}{doi: \begingroup \urlstyle{rm}\Url}\fi

\bibitem[Aan{\ae}s et~al.(2016)Aan{\ae}s, Jensen, Vogiatzis, Tola, and Dahl]{aanaes2016large}
Henrik Aan{\ae}s, Rasmus~Ramsb{\o}l Jensen, George Vogiatzis, Engin Tola, and Anders~Bjorholm Dahl.
\newblock Large-scale data for multiple-view stereopsis.
\newblock \emph{IJCV}, 2016.

\bibitem[Alaluf et~al.(2024)Alaluf, Garibi, Patashnik, Averbuch-Elor, and Cohen-Or]{alaluf2024cross}
Yuval Alaluf, Daniel Garibi, Or Patashnik, Hadar Averbuch-Elor, and Daniel Cohen-Or.
\newblock Cross-image attention for zero-shot appearance transfer.
\newblock In \emph{ACM SIGGRAPH}, 2024.

\bibitem[Brooks et~al.(2023)Brooks, Holynski, and Efros]{Brooks:2023InstructPix2Pix}
Tim Brooks, Aleksander Holynski, and Alexei~A. Efros.
\newblock {InstructPix2Pix}: Learning to follow image editing instructions.
\newblock In \emph{CVPR}, 2023.

\bibitem[Chen et~al.(2022)Chen, Xu, Geiger, Yu, and Su]{Chen2022ECCV}
Anpei Chen, Zexiang Xu, Andreas Geiger, Jingyi Yu, and Hao Su.
\newblock Tensorf: Tensorial radiance fields.
\newblock In \emph{ECCV}, 2022.

\bibitem[Chen et~al.(2023)Chen, Chen, Jiao, and Jia]{chen2023Fantasia3D}
Rui Chen, Yongwei Chen, Ningxin Jiao, and Kui Jia.
\newblock Fantasia{3D}: Disentangling geometry and appearance for high-quality text-to-{3D} content creation.
\newblock In \emph{ICCV}, 2023.

\bibitem[Chou et~al.(2022)Chou, Chugunov, and Heide]{chou2022gensdf}
Gene Chou, Ilya Chugunov, and Felix Heide.
\newblock Gensdf: Two-stage learning of generalizable signed distance functions.
\newblock In \emph{NeurIPS}, 2022.

\bibitem[Chou et~al.(2023)Chou, Bahat, and Heide]{chou2022diffusionsdf}
Gene Chou, Yuval Bahat, and Felix Heide.
\newblock Diffusion-sdf: Conditional generative modeling of signed distance functions.
\newblock 2023.

\bibitem[Deitke et~al.(2022)Deitke, Schwenk, Salvador, Weihs, Michel, VanderBilt, Schmidt, Ehsani, Kembhavi, and Farhadi]{objaverse}
Matt Deitke, Dustin Schwenk, Jordi Salvador, Luca Weihs, Oscar Michel, Eli VanderBilt, Ludwig Schmidt, Kiana Ehsani, Aniruddha Kembhavi, and Ali Farhadi.
\newblock Objaverse: A universe of annotated 3d objects.
\newblock \emph{arXiv preprint arXiv:2212.08051}, 2022.

\bibitem[Deitke et~al.(2023)Deitke, Liu, Wallingford, Ngo, Michel, Kusupati, Fan, Laforte, Voleti, Gadre, VanderBilt, Kembhavi, Vondrick, Gkioxari, Ehsani, Schmidt, and Farhadi]{objaverseXL}
Matt Deitke, Ruoshi Liu, Matthew Wallingford, Huong Ngo, Oscar Michel, Aditya Kusupati, Alan Fan, Christian Laforte, Vikram Voleti, Samir~Yitzhak Gadre, Eli VanderBilt, Aniruddha Kembhavi, Carl Vondrick, Georgia Gkioxari, Kiana Ehsani, Ludwig Schmidt, and Ali Farhadi.
\newblock Objaverse-xl: A universe of 10m+ 3d objects.
\newblock \emph{arXiv preprint arXiv:2307.05663}, 2023.

\bibitem[Erko\c{c} et~al.(2023)Erko\c{c}, Ma, Shan, Nie{\ss}ner, and Dai]{Erkoc_2023_ICCV}
Ziya Erko\c{c}, Fangchang Ma, Qi Shan, Matthias Nie{\ss}ner, and Angela Dai.
\newblock Hyperdiffusion: Generating implicit neural fields with weight-space diffusion.
\newblock In \emph{ICCV}, 2023.

\bibitem[Ghosal et~al.(2023)Ghosal, Majumder, Mehrish, and Poria]{ghosal2023Tango}
Deepanway Ghosal, Navonil Majumder, Ambuj Mehrish, and Soujanya Poria.
\newblock Text-to-audio generation using instruction tuned llm and latent diffusion model.
\newblock \emph{arXiv preprint arXiv:2304.13731}, 2023.

\bibitem[Gropp et~al.(2020)Gropp, Yariv, Haim, Atzmon, and Lipman]{yaron_igr_icml2020}
Amos Gropp, Lior Yariv, Niv Haim, Matan Atzmon, and Yaron Lipman.
\newblock Implicit geometric regularization for learning shapes.
\newblock 2020.

\bibitem[Guo(2022)]{instant-nsr-pl}
Yuan-Chen Guo.
\newblock Instant neural surface reconstruction, 2022.
\newblock https://github.com/bennyguo/instant-nsr-pl.

\bibitem[Guo et~al.(2023)Guo, Liu, Shao, Laforte, Voleti, Luo, Chen, Zou, Wang, Cao, and Zhang]{threestudio2023}
Yuan-Chen Guo, Ying-Tian Liu, Ruizhi Shao, Christian Laforte, Vikram Voleti, Guan Luo, Chia-Hao Chen, Zi-Xin Zou, Chen Wang, Yan-Pei Cao, and Song-Hai Zhang.
\newblock threestudio: A unified framework for 3d content generation.
\newblock \url{https://github.com/threestudio-project/threestudio}, 2023.

\bibitem[Haque et~al.(2023)Haque, Tancik, Efros, Holynski, and Kanazawa]{Haque:2023InstructNeRF}
Ayaan Haque, Matthew Tancik, Alexei Efros, Aleksander Holynski, and Angjoo Kanazawa.
\newblock {Instruct-NeRF2NeRF}: Editing {3D} scenes with instructions.
\newblock In \emph{ICCV}, 2023.

\bibitem[Hertz et~al.(2023)Hertz, Aberman, and Cohen-Or]{hertz2023DDS}
Amir Hertz, Kfir Aberman, and Daniel Cohen-Or.
\newblock Delta denoising score.
\newblock In \emph{ICCV}, 2023.

\bibitem[Ho and Salimans(2021)]{ho2021CFG}
Jonathan Ho and Tim Salimans.
\newblock Classifier-free diffusion guidance.
\newblock In \emph{NeurIPS 2021 Workshop on Deep Generative Models and Downstream Applications}, 2021.

\bibitem[Ho et~al.(2020)Ho, Jain, and Abbeel]{ho2020DDPM}
Jonathan Ho, Ajay Jain, and Pieter Abbeel.
\newblock Denoising diffusion probabilistic models.
\newblock \emph{NeurIPS}, 2020.

\bibitem[Hu et~al.(2022)Hu, Shen, Wallis, Allen-Zhu, Li, Wang, Wang, and Chen]{hu2022lora}
Edward~J Hu, Yelong Shen, Phillip Wallis, Zeyuan Allen-Zhu, Yuanzhi Li, Shean Wang, Lu Wang, and Weizhu Chen.
\newblock Lo{RA}: Low-rank adaptation of large language models.
\newblock In \emph{ICLR}, 2022.

\bibitem[Huang et~al.(2023)Huang, Huang, Yang, Ren, Liu, Li, Ye, Liu, Yin, and Zhao]{huang2023Make-an-Audio}
Rongjie Huang, Jiawei Huang, Dongchao Yang, Yi Ren, Luping Liu, Mingze Li, Zhenhui Ye, Jinglin Liu, Xiang Yin, and Zhou Zhao.
\newblock Make-an-audio: Text-to-audio generation with prompt-enhanced diffusion models.
\newblock \emph{arXiv preprint arXiv:2301.12661}, 2023.

\bibitem[Huberman-Spiegelglas et~al.(2024)Huberman-Spiegelglas, Kulikov, and Michaeli]{huberman2023FriendlyInversion}
Inbar Huberman-Spiegelglas, Vladimir Kulikov, and Tomer Michaeli.
\newblock An edit friendly {DDPM} noise space: Inversion and manipulations.
\newblock In \emph{CVPR}, 2024.

\bibitem[Jun and Nichol(2023)]{jun2023ShapE}
Heewoo Jun and Alex Nichol.
\newblock Shap-e: Generating conditional 3d implicit functions.
\newblock \emph{arXiv preprint arXiv:2305.02463}, 2023.

\bibitem[Kant et~al.(2024)Kant, Siarohin, Wu, Vasilkovsky, Qian, Ren, Guler, Ghanem, Tulyakov, and Gilitschenski]{kant2024spad}
Yash Kant, Aliaksandr Siarohin, Ziyi Wu, Michael Vasilkovsky, Guocheng Qian, Jian Ren, Riza~Alp Guler, Bernard Ghanem, Sergey Tulyakov, and Igor Gilitschenski.
\newblock Spad: Spatially aware multi-view diffusers.
\newblock In \emph{CVPR}, 2024.

\bibitem[Kerbl et~al.(2023)Kerbl, Kopanas, Leimk{\"u}hler, and Drettakis]{kerbl3Dgaussians}
Bernhard Kerbl, Georgios Kopanas, Thomas Leimk{\"u}hler, and George Drettakis.
\newblock 3d gaussian splatting for real-time radiance field rendering.
\newblock \emph{ACM TOG}, 2023.

\bibitem[Koo et~al.(2024)Koo, Park, and Sung]{Koo:2024PDS}
Juil Koo, Chanho Park, and Minhyuk Sung.
\newblock Posterior distillation sampling.
\newblock In \emph{CVPR}, 2024.

\bibitem[Li et~al.(2022)Li, Li, Xiong, and Hoi]{li2022blip}
Junnan Li, Dongxu Li, Caiming Xiong, and Steven Hoi.
\newblock Blip: Bootstrapping language-image pre-training for unified vision-language understanding and generation.
\newblock 2022.

\bibitem[Li et~al.(2023{\natexlab{a}})Li, Gao, Tancik, and Kanazawa]{li2023nerfacc}
Ruilong Li, Hang Gao, Matthew Tancik, and Angjoo Kanazawa.
\newblock Nerfacc: Efficient sampling accelerates nerfs.
\newblock \emph{arXiv preprint arXiv:2305.04966}, 2023{\natexlab{a}}.

\bibitem[Li et~al.(2023{\natexlab{b}})Li, M\"uller, Evans, Taylor, Unberath, Liu, and Lin]{li2023neuralangelo}
Zhaoshuo Li, Thomas M\"uller, Alex Evans, Russell~H Taylor, Mathias Unberath, Ming-Yu Liu, and Chen-Hsuan Lin.
\newblock Neuralangelo: High-fidelity neural surface reconstruction.
\newblock In \emph{CVPR}, 2023{\natexlab{b}}.

\bibitem[Lin et~al.(2023)Lin, Gao, Tang, Takikawa, Zeng, Huang, Kreis, Fidler, Liu, and Lin]{lin2023Magic3D}
Chen-Hsuan Lin, Jun Gao, Luming Tang, Towaki Takikawa, Xiaohui Zeng, Xun Huang, Karsten Kreis, Sanja Fidler, Ming-Yu Liu, and Tsung-Yi Lin.
\newblock Magic{3D}: High-resolution text-to-{3D} content creation.
\newblock In \emph{CVPR}, 2023.

\bibitem[Liu et~al.(2023)Liu, Wu, Van~Hoorick, Tokmakov, Zakharov, and Vondrick]{liu2023zero}
Ruoshi Liu, Rundi Wu, Basile Van~Hoorick, Pavel Tokmakov, Sergey Zakharov, and Carl Vondrick.
\newblock Zero-1-to-3: Zero-shot one image to 3d object.
\newblock In \emph{ICCV}, 2023.

\bibitem[Lombardi et~al.(2019)Lombardi, Simon, Saragih, Schwartz, Lehrmann, and Sheikh]{Lombardi2019TOG}
Stephen Lombardi, Tomas Simon, Jason Saragih, Gabriel Schwartz, Andreas Lehrmann, and Yaser Sheikh.
\newblock Neural volumes: learning dynamic renderable volumes from images.
\newblock \emph{ACM TOG}, 2019.

\bibitem[Meng et~al.(2022)Meng, He, Song, Song, Wu, Zhu, and Ermon]{meng2022SDEdit}
Chenlin Meng, Yutong He, Yang Song, Jiaming Song, Jiajun Wu, Jun-Yan Zhu, and Stefano Ermon.
\newblock {SDE}dit: Guided image synthesis and editing with stochastic differential equations.
\newblock In \emph{ICLR}, 2022.

\bibitem[Metzer et~al.(2023)Metzer, Richardson, Patashnik, Giryes, and Cohen-Or]{metzer2023LatentNeRF}
Gal Metzer, Elad Richardson, Or Patashnik, Raja Giryes, and Daniel Cohen-Or.
\newblock Latent-nerf for shape-guided generation of {3D} shapes and textures.
\newblock In \emph{CVPR}, 2023.

\bibitem[Mildenhall et~al.(2020)Mildenhall, Srinivasan, Tancik, Barron, Ramamoorthi, and Ng]{mildenhall2020NeRF}
Ben Mildenhall, Pratul~P. Srinivasan, Matthew Tancik, Jonathan~T. Barron, Ravi Ramamoorthi, and Ren Ng.
\newblock {NeRF}: Representing scenes as neural radiance fields for view synthesis.
\newblock In \emph{ECCV}, 2020.

\bibitem[M\"uller et~al.(2022)M\"uller, Evans, Schied, and Keller]{mueller2022instant}
Thomas M\"uller, Alex Evans, Christoph Schied, and Alexander Keller.
\newblock Instant neural graphics primitives with a multiresolution hash encoding.
\newblock \emph{ACM TOG}, 2022.

\bibitem[Nam et~al.(2024)Nam, Kwon, Park, and Ye]{Nam_2024_CDS}
Hyelin Nam, Gihyun Kwon, Geon~Yeong Park, and Jong~Chul Ye.
\newblock Contrastive denoising score for text-guided latent diffusion image editing.
\newblock In \emph{CVPR}, 2024.

\bibitem[Nichol et~al.(2022)Nichol, Jun, Dhariwal, Mishkin, and Chen]{nichol2022Point-E}
Alex Nichol, Heewoo Jun, Prafulla Dhariwal, Pamela Mishkin, and Mark Chen.
\newblock Point-e: A system for generating 3d point clouds from complex prompts.
\newblock \emph{arXiv preprint arXiv:2212.08751}, 2022.

\bibitem[Niemeyer et~al.(2020)Niemeyer, Mescheder, Oechsle, and Geiger]{Niemeyer2020CVPR}
Michael Niemeyer, Lars Mescheder, Michael Oechsle, and Andreas Geiger.
\newblock Differentiable volumetric rendering: Learning implicit 3d representations without 3d supervision.
\newblock In \emph{CVPR}, 2020.

\bibitem[Philip and Deschaintre(2023)]{philip2023floaters}
Julien Philip and Valentin Deschaintre.
\newblock Floaters no more: Radiance field gradient scaling for improved near-camera training.
\newblock \emph{Eurographics Symposium on Rendering}, 2023.

\bibitem[Poole et~al.(2023)Poole, Jain, Barron, and Mildenhall]{poole2023DreamFusion}
Ben Poole, Ajay Jain, Jonathan~T. Barron, and Ben Mildenhall.
\newblock Dreamfusion: Text-to-{3D} using {2D} diffusion.
\newblock In \emph{ICLR}, 2023.

\bibitem[Radford et~al.(2021)Radford, Kim, Hallacy, Ramesh, Goh, Agarwal, Sastry, Askell, Mishkin, Clark, et~al.]{radford2021learning}
Alec Radford, Jong~Wook Kim, Chris Hallacy, Aditya Ramesh, Gabriel Goh, Sandhini Agarwal, Girish Sastry, Amanda Askell, Pamela Mishkin, Jack Clark, et~al.
\newblock Learning transferable visual models from natural language supervision.
\newblock 2021.

\bibitem[Raj et~al.(2023)Raj, Kaza, Poole, Niemeyer, Ruiz, Mildenhall, Zada, Aberman, Rubinstein, Barron, et~al.]{raj2023dreambooth3d}
Amit Raj, Srinivas Kaza, Ben Poole, Michael Niemeyer, Nataniel Ruiz, Ben Mildenhall, Shiran Zada, Kfir Aberman, Michael Rubinstein, Jonathan Barron, et~al.
\newblock Dreambooth{3D}: Subject-driven text-to-{3D} generation.
\newblock In \emph{ICCV}, 2023.

\bibitem[Ramesh et~al.(2022)Ramesh, Dhariwal, Nichol, Chu, and Chen]{ramesh2022DALLE2}
Aditya Ramesh, Prafulla Dhariwal, Alex Nichol, Casey Chu, and Mark Chen.
\newblock Hierarchical text-conditional image generation with clip latents.
\newblock \emph{arXiv preprint arXiv:2204.06125}, 2022.

\bibitem[Rombach et~al.(2022)Rombach, Blattmann, Lorenz, Esser, and Ommer]{rombach2022StableDiffusion}
Robin Rombach, Andreas Blattmann, Dominik Lorenz, Patrick Esser, and Bj{\"o}rn Ommer.
\newblock High-resolution image synthesis with latent diffusion models.
\newblock In \emph{CVPR}, 2022.

\bibitem[Ruiz et~al.(2023)Ruiz, Li, Jampani, Pritch, Rubinstein, and Aberman]{ruiz2023DreamBooth}
Nataniel Ruiz, Yuanzhen Li, Varun Jampani, Yael Pritch, Michael Rubinstein, and Kfir Aberman.
\newblock Dreambooth: Fine tuning text-to-image diffusion models for subject-driven generation.
\newblock In \emph{CVPR}, 2023.

\bibitem[RunwayML(2022)]{runwayml-stable-diffusion}
RunwayML.
\newblock Stable diffusion v1.5.
\newblock \url{https://huggingface.co/stable-diffusion-v1-5/stable-diffusion-v1-5}, 2022.

\bibitem[Sch\"{o}nberger and Frahm(2016)]{schoenberger2016sfm}
Johannes~Lutz Sch\"{o}nberger and Jan-Michael Frahm.
\newblock Structure-from-motion revisited.
\newblock In \emph{CVPR}, 2016.

\bibitem[Schuhmann et~al.(2021)Schuhmann, Vencu, Beaumont, Kaczmarczyk, Mullis, Katta, Coombes, Jitsev, and Komatsuzaki]{schuhmann2021laion}
Christoph Schuhmann, Richard Vencu, Romain Beaumont, Robert Kaczmarczyk, Clayton Mullis, Aarush Katta, Theo Coombes, Jenia Jitsev, and Aran Komatsuzaki.
\newblock Laion-400m: Open dataset of clip-filtered 400 million image-text pairs.
\newblock \emph{arXiv preprint arXiv:2111.02114}, 2021.

\bibitem[Schuhmann et~al.(2022)Schuhmann, Beaumont, Vencu, Gordon, Wightman, Cherti, Coombes, Katta, Mullis, Wortsman, et~al.]{schuhmann2022laion}
Christoph Schuhmann, Romain Beaumont, Richard Vencu, Cade Gordon, Ross Wightman, Mehdi Cherti, Theo Coombes, Aarush Katta, Clayton Mullis, Mitchell Wortsman, et~al.
\newblock Laion-5b: An open large-scale dataset for training next generation image-text models.
\newblock In \emph{NeurIPS}, 2022.

\bibitem[Shi et~al.(2023)Shi, Wang, Ye, Long, Li, and Yang]{shi2023MVDream}
Yichun Shi, Peng Wang, Jianglong Ye, Mai Long, Kejie Li, and Xiao Yang.
\newblock {MVD}ream: Multi-view diffusion for {3D} generation.
\newblock \emph{arXiv preprint arXiv:2308.16512}, 2023.

\bibitem[Shim et~al.(2023)Shim, Kang, and Joo]{shim2023diffusion}
Jaehyeok Shim, Changwoo Kang, and Kyungdon Joo.
\newblock Diffusion-based signed distance fields for 3d shape generation.
\newblock In \emph{CVPR}, 2023.

\bibitem[Song et~al.(2021{\natexlab{a}})Song, Meng, and Ermon]{song2021DDIM}
Jiaming Song, Chenlin Meng, and Stefano Ermon.
\newblock Denoising diffusion implicit models.
\newblock In \emph{ICLR}, 2021{\natexlab{a}}.

\bibitem[Song and Ermon(2019)]{song2019NCSN}
Yang Song and Stefano Ermon.
\newblock Generative modeling by estimating gradients of the data distribution.
\newblock In \emph{NeurIPS}, 2019.

\bibitem[Song et~al.(2021{\natexlab{b}})Song, Sohl-Dickstein, Kingma, Kumar, Ermon, and Poole]{song2021SGM}
Yang Song, Jascha Sohl-Dickstein, Diederik~P Kingma, Abhishek Kumar, Stefano Ermon, and Ben Poole.
\newblock Score-based generative modeling through stochastic differential equations.
\newblock In \emph{ICLR}, 2021{\natexlab{b}}.

\bibitem[Sun et~al.(2022)Sun, Sun, and Chen]{SunSC22}
Cheng Sun, Min Sun, and Hwann{-}Tzong Chen.
\newblock Direct voxel grid optimization: Super-fast convergence for radiance fields reconstruction.
\newblock In \emph{CVPR}, 2022.

\bibitem[Tancik et~al.(2023)Tancik, Weber, Ng, Li, Yi, Kerr, Wang, Kristoffersen, Austin, Salahi, Ahuja, McAllister, and Kanazawa]{nerfstudio}
Matthew Tancik, Ethan Weber, Evonne Ng, Ruilong Li, Brent Yi, Justin Kerr, Terrance Wang, Alexander Kristoffersen, Jake Austin, Kamyar Salahi, Abhik Ahuja, David McAllister, and Angjoo Kanazawa.
\newblock Nerfstudio: A modular framework for neural radiance field development.
\newblock In \emph{ACM SIGGRAPH}, 2023.

\bibitem[Vachha and Haque(2024)]{igs2gs}
Cyrus Vachha and Ayaan Haque.
\newblock Instruct-gs2gs: Editing 3d gaussian splats with instructions, 2024.

\bibitem[Wang et~al.(2021)Wang, Liu, Liu, Theobalt, Komura, and Wang]{wang2021neus}
Peng Wang, Lingjie Liu, Yuan Liu, Christian Theobalt, Taku Komura, and Wenping Wang.
\newblock Neus: Learning neural implicit surfaces by volume rendering for multi-view reconstruction.
\newblock In \emph{NeurIPS}, 2021.

\bibitem[Wang et~al.(2023)Wang, Lu, Wang, Bao, Li, Su, and Zhu]{wang2023prolificdreamer}
Zhengyi Wang, Cheng Lu, Yikai Wang, Fan Bao, Chongxuan Li, Hang Su, and Jun Zhu.
\newblock Prolificdreamer: High-fidelity and diverse text-to-3d generation with variational score distillation.
\newblock In \emph{NeurIPS}, 2023.

\bibitem[Wu and la~Torre(2023)]{wu2023CycleDiffusion}
Chen~Henry Wu and Fernando~De la Torre.
\newblock A latent space of stochastic diffusion models for zero-shot image editing and guidance.
\newblock In \emph{ICCV}, 2023.

\bibitem[Xu et~al.(2022)Xu, Xu, Philip, Bi, Shu, Sunkavalli, and Neumann]{xu2022point}
Qiangeng Xu, Zexiang Xu, Julien Philip, Sai Bi, Zhixin Shu, Kalyan Sunkavalli, and Ulrich Neumann.
\newblock Point-nerf: Point-based neural radiance fields.
\newblock In \emph{CVPR}, 2022.

\bibitem[Yang et~al.(2023)Yang, Yu, Wang, Wang, Weng, Zou, and Yu]{yang2023DiffSound}
Dongchao Yang, Jianwei Yu, Helin Wang, Wen Wang, Chao Weng, Yuexian Zou, and Dong Yu.
\newblock Diffsound: Discrete diffusion model for text-to-sound generation.
\newblock \emph{IEEE/ACM Transactions on Audio, Speech, and Language Processing}, 2023.

\bibitem[Yariv et~al.(2020)Yariv, Kasten, Moran, Galun, Atzmon, Ronen, and Lipman]{yariv2020multiview}
Lior Yariv, Yoni Kasten, Dror Moran, Meirav Galun, Matan Atzmon, Basri Ronen, and Yaron Lipman.
\newblock Multiview neural surface reconstruction by disentangling geometry and appearance.
\newblock In \emph{NeurIPS}, 2020.

\bibitem[Yariv et~al.(2021)Yariv, Gu, Kasten, and Lipman]{yariv2021volume}
Lior Yariv, Jiatao Gu, Yoni Kasten, and Yaron Lipman.
\newblock Volume rendering of neural implicit surfaces.
\newblock In \emph{NeurIPS}, 2021.

\bibitem[Ye et~al.(2024)Ye, Li, Kerr, Turkulainen, Yi, Pan, Seiskari, Ye, Hu, Tancik, and Kanazawa]{ye2024gsplatopensourcelibrarygaussian}
Vickie Ye, Ruilong Li, Justin Kerr, Matias Turkulainen, Brent Yi, Zhuoyang Pan, Otto Seiskari, Jianbo Ye, Jeffrey Hu, Matthew Tancik, and Angjoo Kanazawa.
\newblock gsplat: An open-source library for {Gaussian} splatting.
\newblock \emph{arXiv preprint arXiv:2409.06765}, 2024.

\bibitem[Yu et~al.(2021)Yu, Li, Tancik, Li, Ng, and Kanazawa]{yu2021plenoctrees}
Alex Yu, Ruilong Li, Matthew Tancik, Hao Li, Ren Ng, and Angjoo Kanazawa.
\newblock {PlenOctrees} for real-time rendering of neural radiance fields.
\newblock In \emph{ICCV}, 2021.

\bibitem[Yu et~al.(2022)Yu, Chen, Antic, Peng, Bhattacharyya, Niemeyer, Tang, Sattler, and Geiger]{yu2022SDFStudio}
Zehao Yu, Anpei Chen, Bozidar Antic, Songyou Peng, Apratim Bhattacharyya, Michael Niemeyer, Siyu Tang, Torsten Sattler, and Andreas Geiger.
\newblock Sdfstudio: A unified framework for surface reconstruction, 2022.

\bibitem[Zhang et~al.(2020)Zhang, Riegler, Snavely, and Koltun]{zhang2020nerf++}
Kai Zhang, Gernot Riegler, Noah Snavely, and Vladlen Koltun.
\newblock Nerf++: Analyzing and improving neural radiance fields.
\newblock \emph{arXiv preprint arXiv:2010.07492}, 2020.

\bibitem[Zhang et~al.(2023)Zhang, Rao, and Agrawala]{zhang2023adding}
Lvmin Zhang, Anyi Rao, and Maneesh Agrawala.
\newblock Adding conditional control to text-to-image diffusion models, 2023.

\bibitem[Zhang et~al.(2018)Zhang, Isola, Efros, Shechtman, and Wang]{zhang2018perceptual}
Richard Zhang, Phillip Isola, Alexei~A Efros, Eli Shechtman, and Oliver Wang.
\newblock The unreasonable effectiveness of deep features as a perceptual metric.
\newblock In \emph{CVPR}, 2018.

\bibitem[Zhu and Zhuang(2024)]{zhu2023HiFA}
Joseph Zhu and Peiye Zhuang.
\newblock {HiFA}: High-fidelity text-to-{3D} with advanced diffusion guidance.
\newblock In \emph{ICLR}, 2024.

\bibitem[Zhuang et~al.(2023)Zhuang, Wang, Lin, Liu, and Li]{zhuang2023dreameditor}
Jingyu Zhuang, Chen Wang, Liang Lin, Lingjie Liu, and Guanbin Li.
\newblock Dreameditor: Text-driven 3d scene editing with neural fields.
\newblock In \emph{SIGGRAPH Asia}, 2023.

\end{thebibliography}
}
\clearpage
\setcounter{page}{1}

\twocolumn[{%
\renewcommand\twocolumn[1][]{#1}%
\maketitlesupplementary
\includegraphics[width=\linewidth]{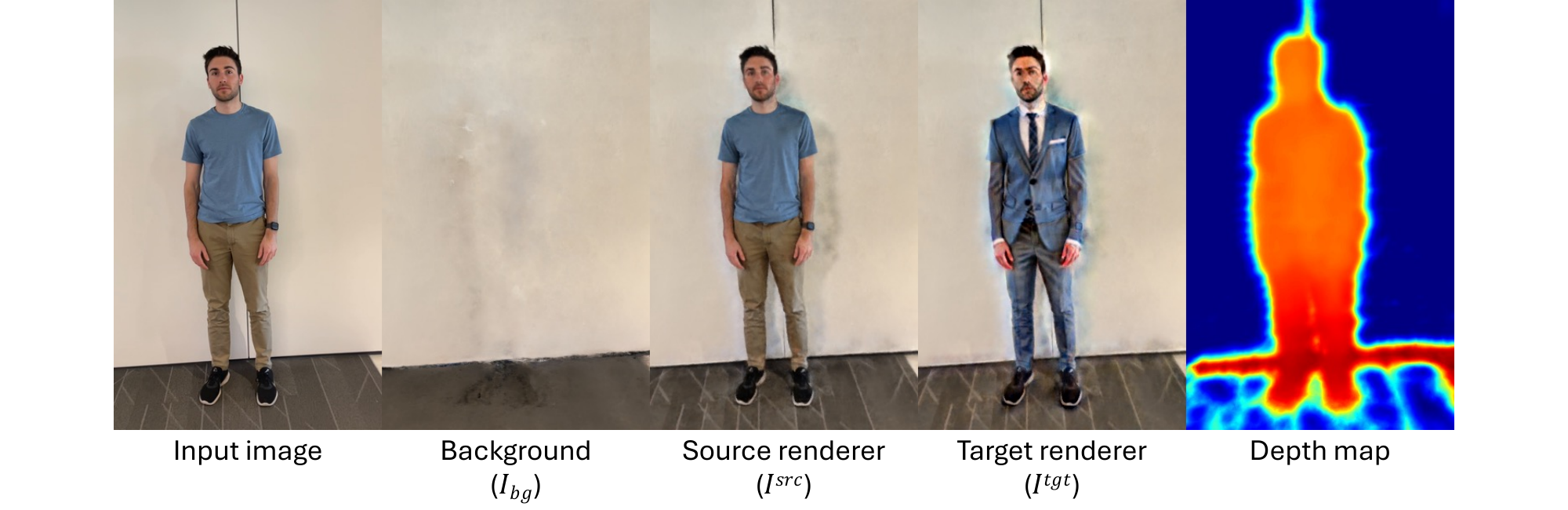}
\captionof{figure}{Our network architecture retains information about the identity of the original scene, including background and foreground details, while simultaneously learning the edited render and geometry. Text prompt: \textit{put him into a suit}
\vspace{2em}}
\label{fig:network_capability_supp}
}]
\section{Network Architecture Details}
\label{sec:arch_details}
\begin{figure*}[!h]
    \centering
\includegraphics[width=\textwidth]{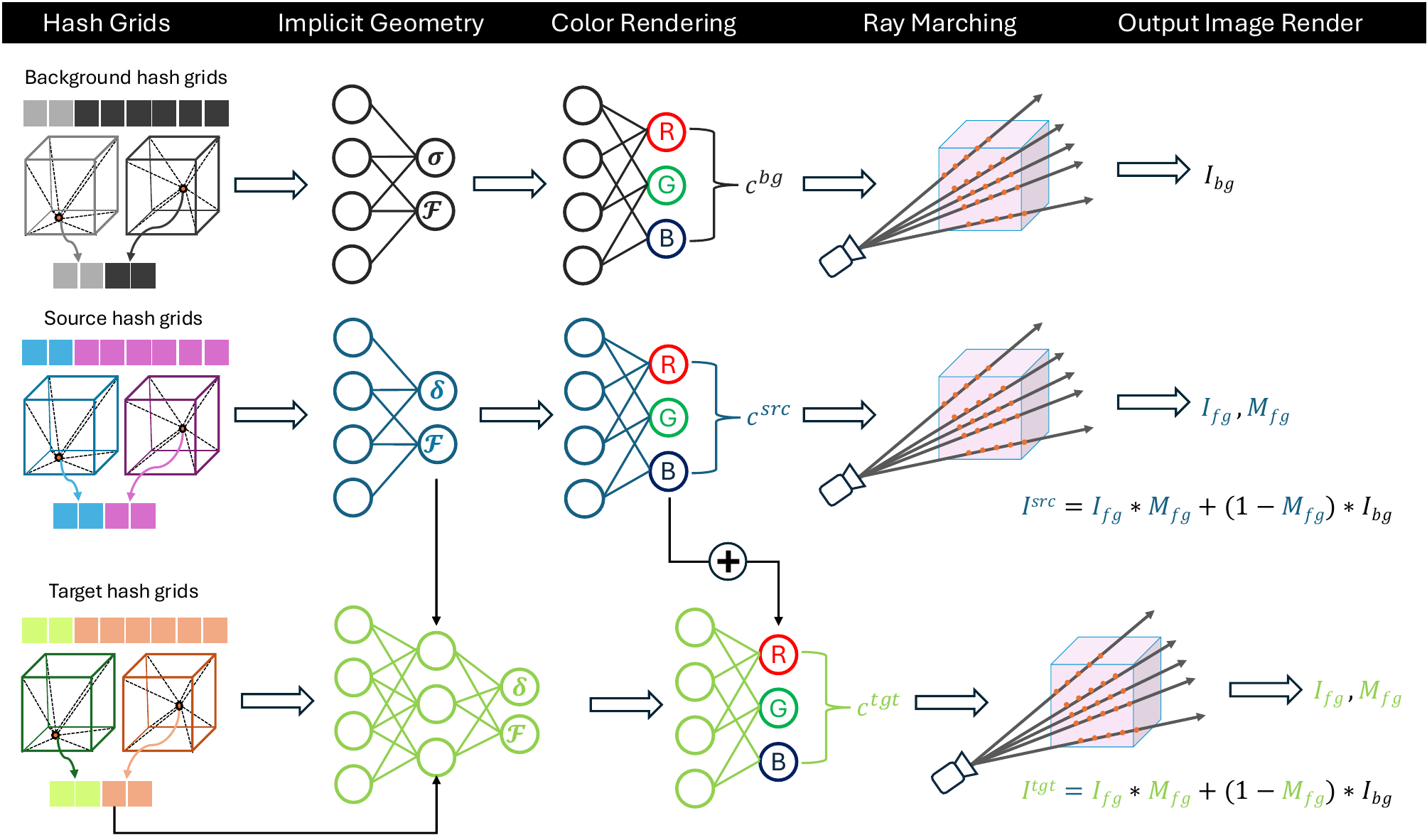}
    \caption{NeuSEditor integrates three dedicated renderers: a background renderer utilizing density fields, source and target foreground renderers employing SDF-based geometry with NeuS volume rendering. The target renderer is initialized from and conditioned on the source to preserve original scene features during editing.}
    \label{fig:renderer}
\end{figure*}

As shown in \cref{fig:network_capability_supp}, our network architecture simultaneously captures three critical components: (1) the background scene context, (2) the source input's foreground elements, and (3) the target edited scene composition.  The architecture comprises three specialized renderers: a background renderer, a source foreground renderer, and a target foreground renderer, as depicted in \cref{fig:renderer}. In this section, we elaborate on the details of each component and their implementation. Identity in this work refers to the composite information captured by the source renderer and background renderer during stage 1 (see \cref{sec:method}).

\paragraph{Background Renderer}
For unbounded scenes, computing a Signed Distance Field (SDF) for the background is impractical; thus, we employ density fields instead. The background renderer operates with its own hash grids and includes subnetworks for implicit geometry and color rendering. The implicit geometry subnetwork processes hash encodings of a spatial point to produce a density value $\sigma$ and a feature vector $\mathcal{F}$. This feature vector, combined with directional information, is then input to the renderer subnetwork to generate color values. To effectively model the unbounded background, we utilize the Inverted Sphere Parameterization \cite{zhang2020nerf++} for volume rendering, yielding the background image $\mathcal{I}_{\text{bg}}$.

\paragraph{Source Renderer}
The source renderer is tasked with learning the foreground of the input scene. It leverages hash grids and subnetworks for implicit geometry and rendering. The implicit geometry subnetwork outputs an SDF value $\delta^{\text{src}}$ and a feature vector $\mathcal{F}$ for a given point encoding, which, along with directional information, are processed by the renderer subnetwork to determine the color. Using the NeuS volume rendering equations \cite{wang2021neus}, we compute the foreground color information $\mathcal{I}_{\text{fg}}^{\text{src}}$ and the foreground mask $\mathcal{M}_{\text{fg}}^{\text{src}}$. The full source image $\mathcal{I}^{\text{src}}$ is then synthesized by combining the foreground and background images with the mask: $\mathcal{I}^{\text{src}} = \mathcal{M}_{\text{fg}}^{\text{src}} \cdot \mathcal{I}_{\text{fg}}^{\text{src}} + (1 - \mathcal{M}_{\text{fg}}^{\text{src}}) \cdot \mathcal{I}_{\text{bg}}$.

\paragraph{Target Renderer}
The target renderer, designed for the edited scene, mirrors the source renderer’s structure, employing hash grids, an SDF-based implicit geometry subnetwork, and a color renderer subnetwork. It also adopts the NeuS volume rendering equations to derive the foreground image $\mathcal{I}_{\text{fg}}^{\text{tgt}}$ and mask $\mathcal{M}_{\text{fg}}^{\text{tgt}}$ for the target scene. The full target image is composed as $\mathcal{I}^{\text{tgt}} = \mathcal{M}_{\text{fg}}^{\text{tgt}} \cdot \mathcal{I}_{\text{fg}}^{\text{tgt}} + (1 - \mathcal{M}_{\text{fg}}^{\text{tgt}}) \cdot \mathcal{I}_{\text{bg}}$. To preserve recognizable features from the original scene during the editing (see \cref{sec:method}), the target renderer is initialized to approximate the source rendering and geometry. The target geometry network is conditioned on the source geometry network, guiding it to generate a geometry (semantically) proximate to the source. Meanwhile, the color network focuses on tuning and editing the source scene’s colors while retaining its access to source colors.

By integrating these three renderers, our architecture facilitates the cohesive modeling of the background, the original foreground, and the edited foreground, enabling seamless scene editing.

\begin{figure*}[!h]
    \centering
    \includegraphics[width=\textwidth]{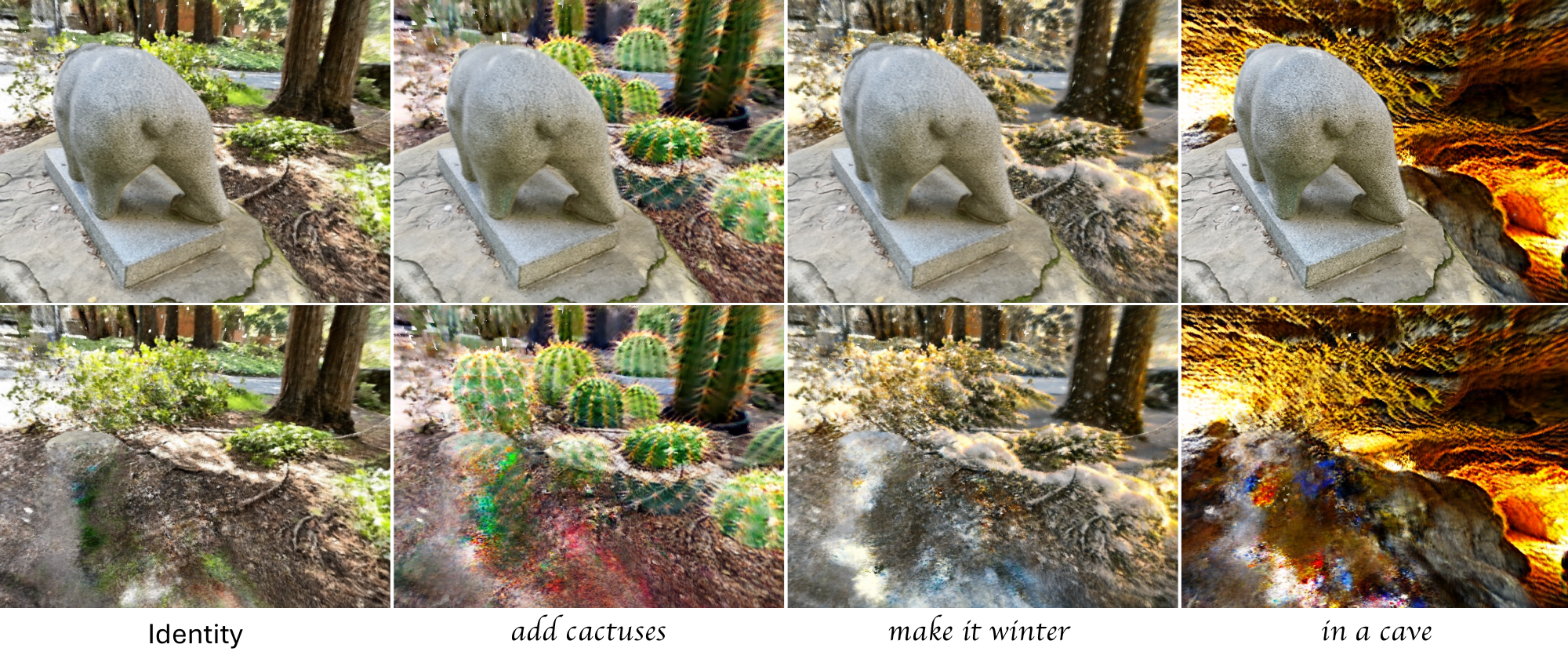}
    
    \vspace{5pt} % Adds some space before the line
    \hrule height 1pt
    \vspace{5pt} % Adds some space after the line
    
    \includegraphics[width=\textwidth]{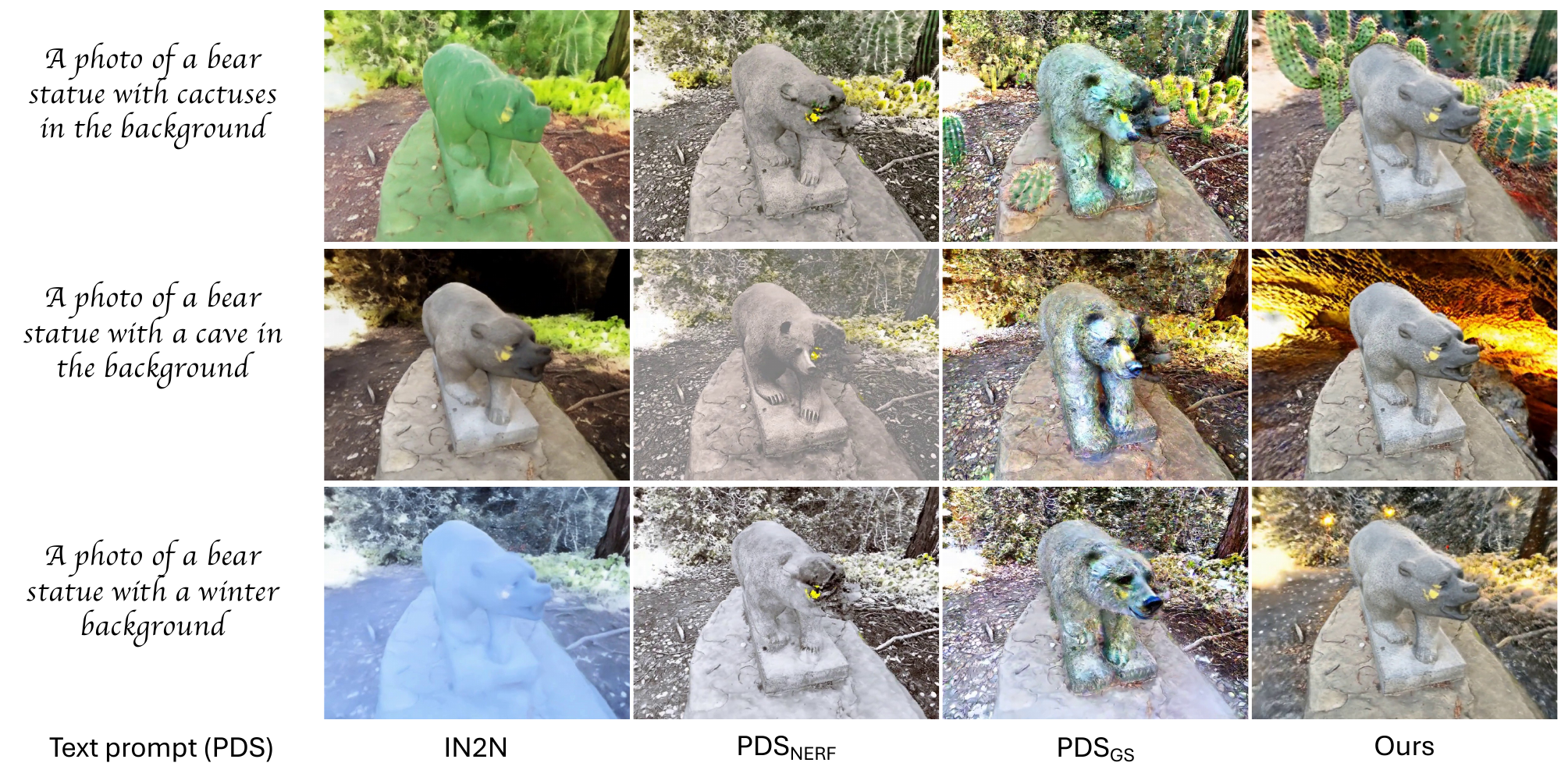}
    \caption{Background editing results. Top row shows the full model render, while the bottom row displays the edited background under various text prompts. Our method effectively modifies the background while preserving the foreground integrity, outperforming competing methods that alter foreground details.}
    \label{fig:background_comparison}
\end{figure*}

\section{Background editing }
\label{sec:background_editting}
As discussed in \cref{sec:arch_details}, the background is modeled using radiance fields, with a separate set of hash grids as positional encodings. This background model utilizes its own geometry and renderer subnetworks. 
Our method also supports background editing.  Similar to foreground editing, we still need to learn the input scene first to decompose it into foreground and background elements.

Unlike foreground editing, the use of ``additive learning'' techniques is less crucial for background editing. 
%This is because background modifications are generally more generative in nature, aiming to completely change the scene rather than adjusting specific elements. 
This is because background modifications are generally more generative in nature, aiming to completely change the scene rather than adjusting specific elements.
Thus, to reduce computational demands, we directly edit the parameters of the background renderer instead of relying on ``additive learning''.

\cref{fig:background_comparison} shows the background editing capabilities of our approach. The top section displays two rows of images: the upper row shows the complete model render, while the lower row presents the isolated background render under different text prompts. These results demonstrate that our method can effectively modify the background while preserving the integrity of the foreground. The lower section compares our method with three recent approaches. Notably, despite explicit background specifications and various prompt engineering attempts, competing methods persistently modify foreground elements. In contrast, our approach preserves the foreground by exclusively optimizing background parameters. Thus, the explicit separation of background and foreground renderers further enhances our control over edits.
\begin{table}[h]
    \centering
    \begin{tabular}{lcccccc}
        \hline
        Metric & Backbone & IN2N & PDS\textsubscript{NERF} & PDS\textsubscript{GS} & Ours \\
        \hline
        \multirow{2}{*}{clip $\uparrow$} & ViT/16 & 0.317 & 0.318 & \cellcolor{red!25}0.336 & \cellcolor{yellow!25}0.334 \\
                             & ViT/32 & 0.318 & 0.312 & \cellcolor{red!25}0.341 & \cellcolor{yellow!25}0.319 \\
        \hline
        \multirow{2}{*}{lpips $\downarrow$} & Alex   & 0.748 & \cellcolor{yellow!25}0.668 & 0.737 & \cellcolor{red!25}0.533 \\
                               & VGG    & 0.697 & \cellcolor{yellow!25}0.615 & 0.658 & \cellcolor{red!25}0.500 \\
        \hline
    \end{tabular}
\caption{The top row shows text-image similarity, and the bottom shows perceptual similarity between the input and edited scenes.}
    \label{tab:background_comparison}
\end{table}

To quantitatively evaluate our background editing, we utilize both CLIP and LPIPS metrics. The CLIP score measures text-render alignment, while the LPIPS distance assesses the preservation of foreground details.  To ensure an unbiased evaluation of the LPIPS distance, we trained a separate model solely to render the foreground components of the scene. Ideally, a high CLIP score combined with a low LPIPS distance indicates that modifications are restricted to the background. As shown in \cref{tab:background_comparison}, our method achieves competitive text-image alignment while maintaining a lower LPIPS distance compared to other approaches. These results confirm that our edits are effectively confined to the background, contributing positively to the overall text-image alignment.

\section{Avoiding mode collapse}
\label{sec:mode_collapse}
\begin{figure*}[h!]
    \centering
\includegraphics[width=\textwidth]{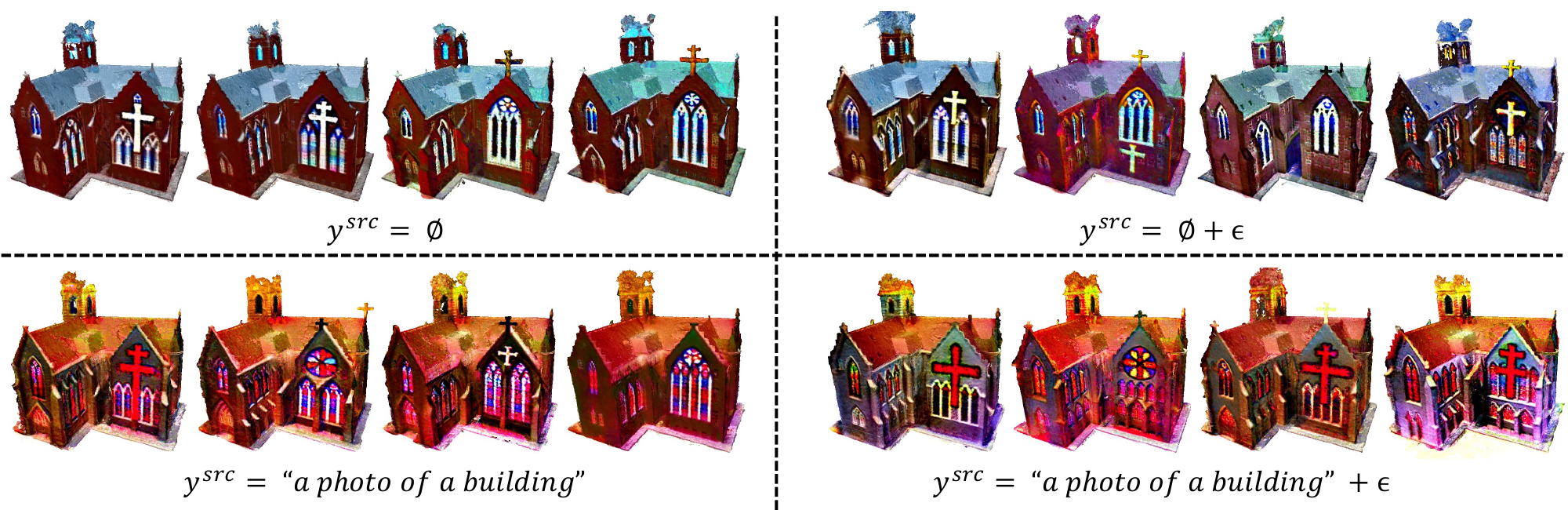}
    \caption{This figure shows textured mesh reconstruction results under varying text prompts. The top-left quadrant depicts results without a source prompt (\textit{null} prompt), the top-right quadrant illustrates results with a \textit{null} prompt and slight perturbations to the source text embeddings. The bottom-left quadrant demonstrates results with a descriptive source prompt, while the bottom-right quadrant presents results where small perturbations are applied to the same descriptive source text embeddings.}
    \label{fig:dist_coverage}
\end{figure*}

Mode collapse is a common issue in generative AI pipelines (e.g. GANs, GPTs, text-to-3Ds) where the generator learns to produce a limited set of (or similar) outputs, ignoring the full diversity of the target data distribution. 
Attentive readers may have observed from \cref{fig:teaser,fig:ablation} that our method is capable of generating diverse edits using the same prompt (``\textit{make it a church'}'). However, like most generative techniques, text-guided editing approaches are prone to mode collapse. To mitigate this issue, we propose two simple yet effective techniques that leverage source prompting.

\cref{eq:source_prompt} shows that our model can consistently apply edits with or without the source prompt. However, including the source prompt allows users to guide the editing process towards exploring different modes of the distribution.
Additionally, to improve distribution coverage, we introduce a straightforward technique that involves injecting a small amount of noise into the source prompt embedding. By utilizing this noise, users can steer the network to generate varied edits that align closely with the target text.

\cref{fig:dist_coverage} provides qualitative evidence supporting the effectiveness of our proposed strategies for improving distribution coverage. The figure demonstrates that appending a descriptive source prompt effectively shifts the mode of the distribution. Furthermore, introducing a slight perturbation to the embedding of the source prompt encourages greater diversity in the reconstructed outputs.
\begin{figure}[h!]
    \centering
    \includegraphics[width=\columnwidth]{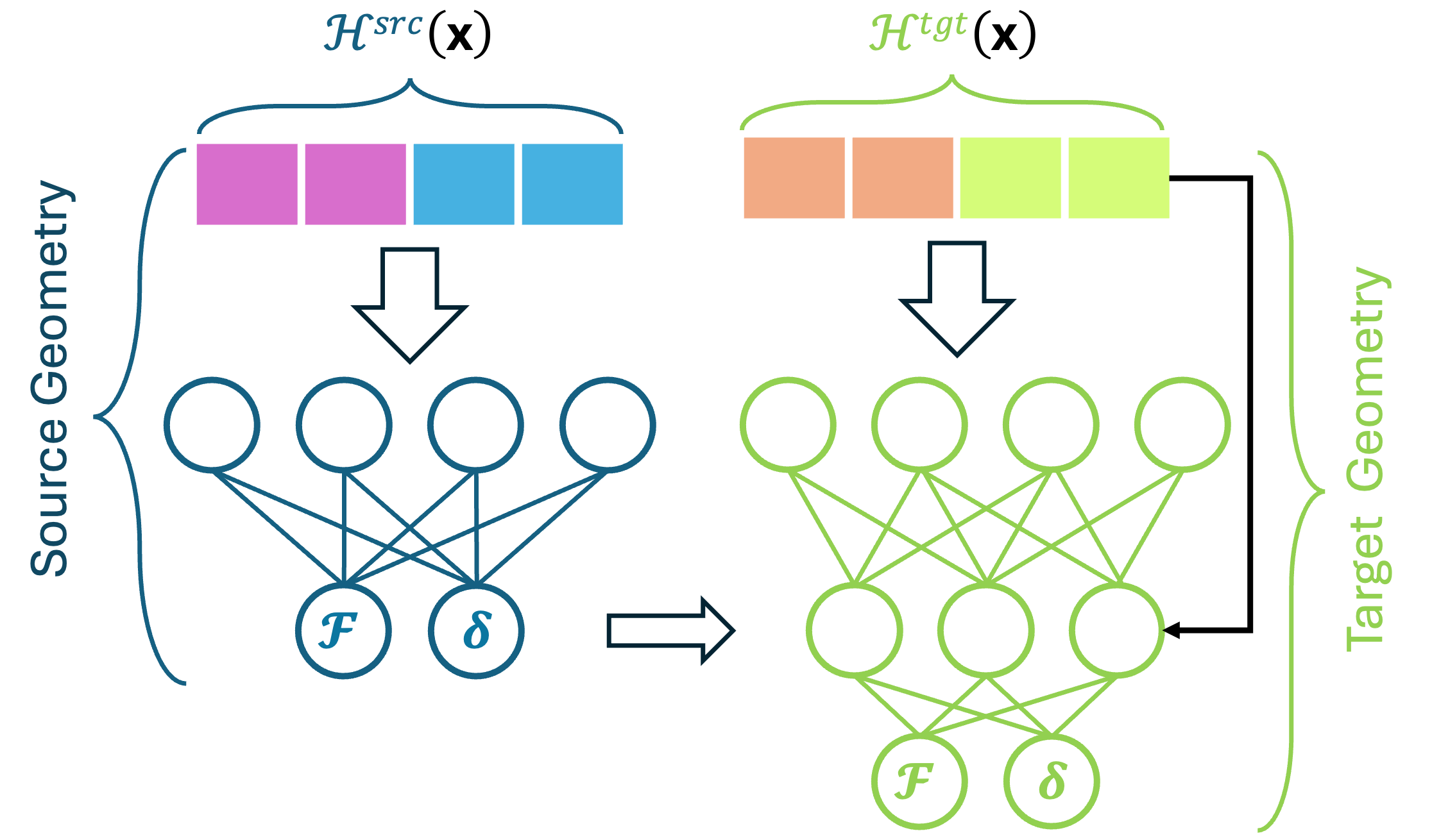}
    \caption{Source and target geometry networks.}
    \label{fig:sdf_grad}
\end{figure}
\begin{figure*}[]
\centering
\includegraphics[width=\textwidth]{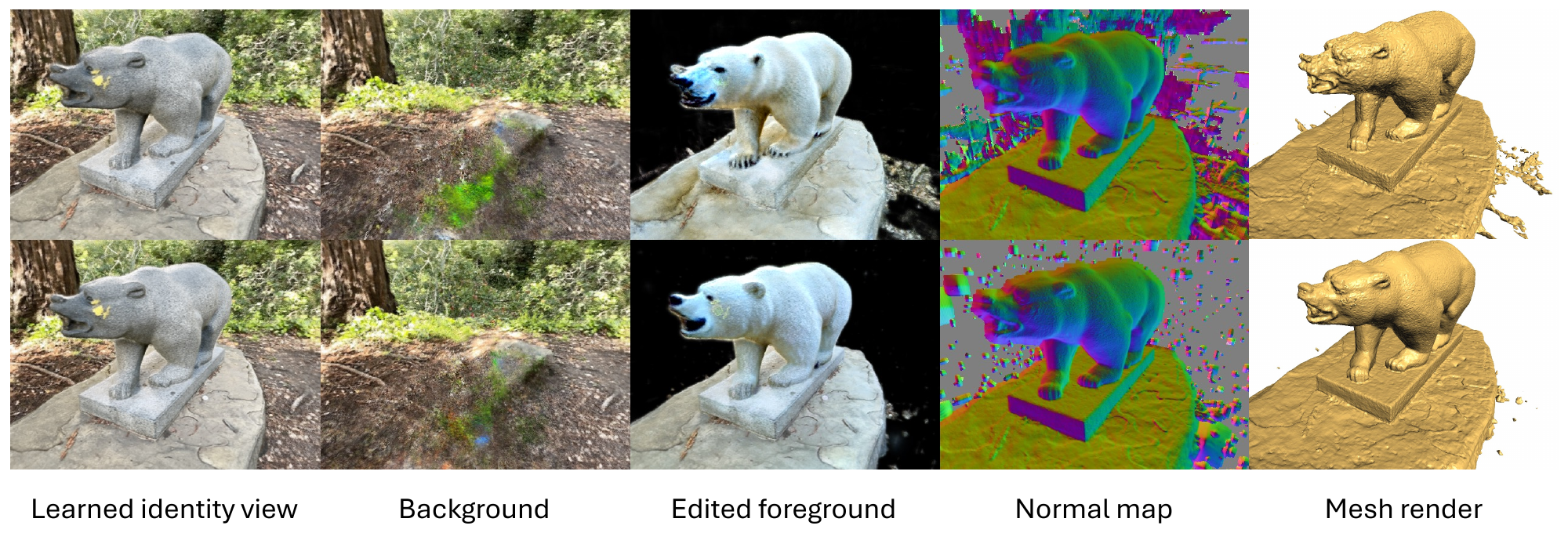}
    \caption{Experiments conducted on the IN2N dataset (bear scene) demonstrate both analytical and numerical solutions for SDF gradient computations. From left to right, the figure demonstrates the full identity rendering, background rendering, edited foreground rendering, normalized SDF gradient output from the network represented as a normal map, and the corresponding mesh rendering. The top row shows SDF gradients computed using the analytical solution, while the bottom row utilizes the numerical solution.}
    \label{fig:gradient_type}
\end{figure*}
\section{SDF gradient computation}
%Talk about how numerical SDF gradient works better than numerical
We have observed that, during editing, the numerical computation of the SDF gradient (via \textit{finite differences}) results in cleaner and smoother geometry compared to using the analytical gradient via \textit{torch.autograd.grad}. \cref{fig:sdf_grad} shows the geometry networks for both the source and target. These networks process 3D points using hash encodings combined with MLPs,  outputting the SDFs and corresponding geometric feature vectors for the source and target scenes.
The \textit{true} gradient for the  source (identity) SDF can be computed using the following equations:
\begin{equation}
   \delta^{\text{src}}(\mathbf{x}), \mathcal{F}^{\text{src}}(\mathbf{x}) = \mathbf{G}^{\text{src}}(\mathbf{x}) = \mathrm{MLP}^{\text{src}}(\mathcal{H}^{\text{src}}(\mathbf{x})) 
\end{equation}
\begin{equation}
    \nabla \delta^{\text{src}}(\mathbf{x})=  \lim_{h \to 0}\frac{\delta^{\text{src}}(\mathbf{x}+h) - \delta^{\text{src}}(\mathbf{x}-h)}{2h}
\end{equation}
 where \(\mathbf{x}\) is the 3D query point position, \(\mathcal{H}^{\text{src}}\) represents the hash encodings of the query point, \(\mathbf{G}^{\text{src}}\) denotes the MLP-based geometry module for the source scene, and \(\delta^{\text{src}}\) and \(\mathcal{F}^{\text{src}}\) are the respective output SDF and features from the geometry module.
 As depicted in \cref{fig:sdf_grad}, the target geometry is conditioned on the outputs of the source geometry network.
 Thus, the  solution for the \textit{true} target (edited)  SDF gradient can be computed as follows:
\begin{equation}
   \delta^{\text{tgt}}(\mathbf{x}), \mathcal{F}^{\text{tgt}}(\mathbf{x}) = \mathrm{MLP}^{\text{tgt}}(\mathbf{G}^{\text{src}}(\mathbf{x}), \mathcal{H}^{\text{tgt}}(\mathbf{x}))
\end{equation}
\begin{equation}
    \nabla \delta^{\text{tgt}}(\mathbf{x}) =  \lim_{h \to 0} \frac{\delta^{\text{tgt}}(\mathbf{x}+h) - \delta^{\text{tgt}}(\mathbf{x}-h)}{2h}
\end{equation}
where \(\mathbf{x}\) is the 3D query point position, \(\mathcal{H}^{\text{tgt}}\) represents the target hash encodings of the query point, \(\mathrm{MLP}^{\text{tgt}}\) is the MLP-based feature and geometry extraction module for the target scene, which utilizes knowledge from the source geometry. The terms \(\delta^{\text{tgt}}\) and \(\mathcal{F}^{\text{tgt}}\) represent the respective output SDF and features of the target scene. 
For both the source and target scenes, we \textit{numerically} approximate the \textit{true} gradient by setting \( h \) to a very small positive value.

\begin{table}[h]
    \centering
    \begin{tabular}{lccc}
        \hline
        Metric & Backbone & Analytical & Numerical \\
        \hline
        \multirow{2}{*}{clip $\uparrow$} & ViT/16  & \textbf{0.298} & 0.294  \\
                             & ViT/32  & \textbf{0.290} & 0.285  \\
        \hline
        \multirow{2}{*}{lpips $\downarrow$} & Alex    & 0.292 & \textbf{0.201}  \\
                               & VGG     & 0.353 &  \textbf{0.262}  \\
        \hline
    \end{tabular}
\caption{Gradient type evaluation using CLIP and LPIPS.}
    \label{tab:sdf_grad}
\end{table}

Our experiments show that numerical SDF gradients outperform analytical gradients in capturing target geometry.
\cref{fig:gradient_type} illustrates experiments with a bear scene prompted with “\textit{turn the bear into a polar bear}”. The numerical SDF gradient computation (bottom row), representing the output of the fusion module in ~\cref{fig:sdf_grad}, significantly enhances the network's ability to capture foreground details compared to the analytical gradient (top row) during the editing process. 
The normal map in the \cref{fig:gradient_type} is computed using the SDF gradient output of the fusion module.
For mesh extraction, we utilized the marching cubes algorithm, and our experiments revealed that numerical gradients also contribute to cleaner surface reconstructions.

We quantitatively evaluate the  editing quality of the experiment. \cref{tab:sdf_grad} presents the evaluation of different gradient types for this 3D editing task. Our results indicate that both gradient types yield similar CLIP metrics, suggesting that the choice of gradient type does not significantly affect text image (target rendering) alignment. In contrast, the LPIPS distance is notably reduced when using numerical gradients compared to analytical gradients. As qualitative observations also suggest, we believe that this improvement is due to numerical gradients generating fewer floaters and yielding smoother geometric surfaces.
\section{Dataset and benchmark details}
\label{sec:user_survey_details}
\begin{algorithm}[H]
\small
\caption{Generate DTU Spherical Camera Poses}
\begin{algorithmic}[1]
\Function{DTUSphericPoses}{cams, n\_steps}
    \State center $\gets [0, 0, 0]$
    \State cam\_center $\gets \text{mean}(\text{cams})$
    \State eigvecs $\gets \text{eigenvectors}(\text{cams}^T \times \text{cams})$
    \State up $\gets \text{eigvecs}[:, 1]$
    \State rot\_dir $\gets \text{cross}(\text{up}, \text{cam\_center})$
    \State max\_angle $\gets \max(\arccos(\text{cams},\text{cam\_center}))$
    \State poses $\gets [\ ]$
    \For{$\theta$ in $\text{linspace}(-\text{max\_angle}, \text{max\_angle}, \text{n\_steps})$}
        \State cam\_pos $\gets \text{cam\_center} \cdot \cos(\theta) + \text{rot\_dir} \cdot \sin(\theta)$
        \State look\_dir $\gets \text{(center - cam\_pos)}.norm()$
        \State side $\gets (\text{look\_dir} \times \text{up}).norm()$
        \State up\_vec $\gets (\text{side} \times \text{look\_dir}).norm()$
        \State pose $\gets [\text{side}, \text{up\_vec}, -\text{look\_dir}, \text{cam\_pos}]$ \#SE(3) 
        \State Append pose to poses
    \EndFor
    \State \Return poses
\EndFunction
\end{algorithmic}
\label{alg:spherical_pose}
\end{algorithm}

\paragraph{Spherical rendering setup.} 
As discussed in \cref{sec:dataset}, we rendered videos using spherical poses for the DTU scenes and the IN2N \textit{bear} scene to ensure a fair evaluation. The DTU dataset contains scenes with 49 or 64 images, where the first 49 images are distributed on the same upper hemisphere. To compute the spherical rendering path of the DTU scene, we calculate the average distance from the camera poses to the center  $(0., 0., 0.)$ and set it as the radius. Additionally, we assumed that the second eigenvector of the camera positions aligns with the vertical ``up'' vector, given the predominant horizontal and (then) vertical distribution of the cameras. Full details are provided in \cref{alg:spherical_pose}.
For the \textit{bear} scene, we followed a similar approach, leveraging colmap SfM~\cite{schoenberger2016sfm} data. Here, we assumed the ``up'' vector to be normal of the dominant ground plane, with the scene center defined as the closest ``intersection'' point of the cameras' ``look-at'' rays.

\paragraph{Finetuning of IN2N and PDS methods.}
We devoted significant effort to finetuning the IN2N and PDS methods to improve their results. Since these methods rely on nerfstudio~\cite{nerfstudio}, we ensured that camera optimizers were disabled, as we primarily used groundtruth poses. Additionally, all poses were included in the training set, with no views excluded for validation. 

IN2N was found to be sensitive to hyperparameters and prone to catastrophic forgetting. To address this in Blender and DTU scenes, we trained each scene using three hyperparameter configurations: (1) default settings with \textit{text guidance} of 7.5 and \textit{image guidance} of 1.5; (2) increased \textit{image guidance} of 2.5; and (3) reduced \textit{text guidance} of 5.0. The best results were selected, with default settings used as a fallback in case all of them leading to degenerate cases. For IN2N-data, we used the hyperparameter values specified in its supplementary material and followed recommendations from GitHub discussions (\href{https://github.com/ayaanzhaque/instruct-nerf2nerf/issues/60}{1}, \href{https://github.com/ayaanzhaque/instruct-nerf2nerf/issues/101}{2}).

For PDS methods, we converted all datasets to the \textit{nerfstudio-data} format, as their dataparser requires this. Each experiment was run multiple times, and the best result was selected. PDS\textsubscript{NeRF} was more prone to degenerate cases with monotonic colors. In cases where PDS\textsubscript{NeRF} led to degenerate outputs, early stopping was considered.

For our method, we utilized analytical SDF gradients and a shorter identity learning phase (8K iterations at stage 1) in DTU and Blender scenes to better demonstrate our architecture's identity-preserving capabilities, whereas other methods were trained 30K iterations for identity learning.

\paragraph{User study results details.}
\cref{tab:survey_details} shows the user study results for each experiment.
In the DTU scenes, our method and PDS\textsubscript{Splat} clearly outperform the other methods, while IN2N struggles with optimization, often leading to degenerate cases. PDS\textsubscript{NeRF} produces excessive floaters, likely leading users to prefer our method and PDS\textsubscript{Splat}.

In the Blender dataset, particularly in the \textit{hotdog} and \textit{mic} scenes, our method and PDS\textsubscript{Splat} again achieve superior performance, whereas PDS\textsubscript{NeRF} struggles due to excessive floater generation. However, IN2N performs better than PDS\textsubscript{NeRF} in these cases, likely because of its lower floater count. In the \textit{ficus} scene, PDS\textsubscript{Splat} fails due to multi-view inconsistencies, IN2N struggles with producing the expected edits, and PDS\textsubscript{NeRF} performs best, aligning more closely with the text prompts.

In experiments using IN2N-data, IN2N achieves the best overall results, as expected, since it utilizes the same prompts, datasets, and hyperparameters found effective in the original paper and supplementary materials. In the \textit{person} scene, all methods generate clean renders compared to previous scenes. However, users prefer the results from PDS\textsubscript{Splat} and IN2N, likely due to their cleaner outputs and better preservation of human identity. In the \textit{bear} scene, our method and IN2N produce comparable results, with users slightly favoring our method, likely because it is more effective at avoiding the ``Janus artifact''. PDS\textsubscript{Splat} suffers from multi-view inconsistencies, while PDS\textsubscript{NeRF} produces monotonic colors.

\begin{table*}[ht]
\centering
\begin{adjustbox}{width=0.85\textwidth}
\begin{tabular}{rllcccc}
\toprule
\# & \textbf{Dataset-Scene} & \textbf{Keyword} & \textbf{IN2N} & \textbf{PDS$_{\text{NeRF}}$} & \textbf{PDS$_{\text{Splat}}$} & \textbf{Ours} \\
\midrule
1 & \multirow{6}{*}{DTU - Scan24} & Church        & 0.20 & 1.37 & \cellcolor{yellow!25}1.68 & \cellcolor{red!25}2.76\\
2 & &                               Mosque        & 0.07 & 1.22 & \cellcolor{yellow!25}2.10 & \cellcolor{red!25}2.61 \\
3 & &                               Castle        & 0.20 & 1.54 & \cellcolor{red!25}2.22 & \cellcolor{yellow!25}2.05 \\
4 & &                               Disney castle & 0.39 & 0.76 & \cellcolor{yellow!25}2.17 & \cellcolor{red!25}2.68 \\
5 & &                               Lego          & 0.15 & 1.37 & \cellcolor{yellow!25}1.90 & \cellcolor{red!25}2.59 \\
6 & &                               Barn          & 0.07 & 1.34 & \cellcolor{yellow!25}1.73 & \cellcolor{red!25}2.85 \\
\midrule
7 & \multirow{5}{*}{DTU - Scan65} & Moustache$^*$ & \cellcolor{yellow!25}1.73 & 0.71 & 0.90 & \cellcolor{red!25}2.66 \\
8 &                               & Horned skull  & 0.17 & 1.00 & \cellcolor{yellow!25}2.02 & \cellcolor{red!25}2.80 \\
9 &                               & Alien         & 0.63 & 0.71 & \cellcolor{yellow!25}1.98 & \cellcolor{red!25}2.68 \\
10 &                              & Buddha        & 0.10 & 1.22 & \cellcolor{yellow!25}1.90 & \cellcolor{red!25}2.78 \\
11 &                              & Clown$^*$     & \cellcolor{yellow!25}1.98 & 0.39 & 1.10 & \cellcolor{red!25}2.54 \\
\midrule
12 & \multirow{2}{*}{DTU - Scan83} & Suit          & 0.07 & \cellcolor{yellow!25}1.88 & 1.66 & \cellcolor{red!25}2.39 \\
13 &                               & Bowtie$^*$    & 0.07 & \cellcolor{yellow!25}2.02 & 1.85 & \cellcolor{red!25}2.05 \\
\midrule
14 & \multirow{3}{*}{DTU - Scan105} & Suit         & 0.46 & 0.66 & \cellcolor{yellow!25}2.12 & \cellcolor{red!25}2.76 \\
15 &                                & Tiger        & 0.63 & \cellcolor{yellow!25}1.83 & 1.59 & \cellcolor{red!25}1.95 \\
16 &                                & Bowtie$^*$   & 0.56 & 1.20 & \cellcolor{yellow!25}1.71 & \cellcolor{red!25}2.54 \\
\midrule
17 & \multirow{2}{*}{DTU - Scan106} & Chickens     & 0.98 & 0.88 & \cellcolor{yellow!25}1.59 & \cellcolor{red!25}2.56 \\
18 &                                & Crows        & 0.07 & 1.49 & \cellcolor{yellow!25}1.76 & \cellcolor{red!25}2.68 \\
\midrule
19 & \multirow{3}{*}{DTU - Scan110}  & Monk             & 0.88 & 0.22 & \cellcolor{yellow!25}2.34 & \cellcolor{red!25}2.56 \\
20 &                                 & Buddha           & 0.98 & 0.12 & \cellcolor{red!25}2.68 & \cellcolor{yellow!25}2.22 \\
21 &                                 & Snoop Dogg$^*$   & 0.63 & 0.56 & \cellcolor{yellow!25}2.20 & \cellcolor{red!25}2.61 \\
\midrule
22 & \multirow{2}{*}{Blender - Hotdog} & Bananas     & 1.27 & 0.22 & \cellcolor{yellow!25}1.85 & \cellcolor{red!25}2.66  \\
23 &                                   & Corns       & 0.90 & 0.56 & \cellcolor{yellow!25}1.68 & \cellcolor{red!25}2.85  \\
\midrule
24 & \multirow{2}{*}{Blender - Mic} & Hair dryer     & 1.34 & 0.27 & \cellcolor{yellow!25}1.93 & \cellcolor{red!25}2.46  \\
25 & & Pistol         & 1.22 & 0.07 & \cellcolor{yellow!25}2.22 & \cellcolor{red!25}2.49 \\
\midrule
26 & \multirow{3}{*}{Blender - Ficus} & Cactus     & 0.24 & 0.98 & \cellcolor{yellow!25}1.85 & \cellcolor{red!25}2.93 \\
27 & & Apple tree         & \cellcolor{yellow!25}1.98 & 1.37 & 0.07 & \cellcolor{red!25}2.59 \\
28 & & Rose bush  & 0.59 & \cellcolor{yellow!25}1.51 & 1.07 & \cellcolor{red!25}2.83  \\
\midrule
29 & \multirow{3}{*}{IN2N - Person} & Clown     & \cellcolor{yellow!25}2.05 & 0.15 & \cellcolor{red!25}2.46 & 1.34 \\
30 & & Suit         & \cellcolor{red!25}2.76 & 0.73 & \cellcolor{yellow!25}1.95 & 0.56 \\
31 & & Firefighter & \cellcolor{red!25}2.29 & 1.07 & \cellcolor{yellow!25}2.17 & 0.46 \\
\midrule
32 & \multirow{3}{*}{IN2N - Bear} & Grizzly bear     & \cellcolor{red!25}2.71 & 0.80 & 0.41 & \cellcolor{yellow!25}2.07 \\
33 & & Panda         & \cellcolor{yellow!25}2.22 & 0.56 & 0.66 & \cellcolor{red!25}2.56 \\
34 & & Polar bear & \cellcolor{yellow!25}2.24 & 0.44 & 0.66 & \cellcolor{red!25}2.66 \\
\bottomrule
\end{tabular}
\end{adjustbox}
\caption{User survey results across 34 experiments. NeuSEditor used a guidance scale of 350 for all experiments, except those marked with an asterisk (*), where a lower scale (100) was applied to encourage minimal changes (e.g. add a moustache/bowtie). We invite readers to compare these quantitative user survey results with the qualitative results shared on the \href{https://neuseditor.github.io/survey_experiments/survey_index.html}{survey page replica}.}
\label{tab:survey_details}
\end{table*}
\end{document}